\documentclass{article} 
\usepackage{iclr2025_conference,times}

\usepackage{amsfonts}       
\usepackage{amsmath}
\usepackage{amssymb}
\usepackage{booktabs}       
\usepackage{float}
\usepackage{graphicx}       
\usepackage{hyperref}       
\usepackage{microtype}      
\usepackage{multirow}
\usepackage{nicefrac}       
\usepackage{pifont}         
\usepackage{url}            
\usepackage{tabularx}
\usepackage{wrapfig}
\usepackage{xcolor}         

\usepackage{cleveref} 

\iclrfinalcopy

\usepackage{amsmath,amsfonts,bm}

\def\eqref#1{equation~\ref{#1}}

\def\1{\bm{1}}

\def\eps{{\epsilon}}

\DeclareMathAlphabet{\mathsfit}{\encodingdefault}{\sfdefault}{m}{sl}
\SetMathAlphabet{\mathsfit}{bold}{\encodingdefault}{\sfdefault}{bx}{n}

\usepackage{hyperref}
\usepackage{url}

\title{Contextual Document Embeddings}

\author{John X. Morris \\
Cornell University \\ 
jxm3@cornell.edu
\And
Alexander M. Rush \\
Cornell University \\ 
arush@cornell.edu
}

\newcommand{\Jack}[1]{}

\newcommand{\ModelName}{\texttt{cde-small-v1}}

\begin{document}

\maketitle

\begin{abstract}
Dense document embeddings are central to neural retrieval. The dominant paradigm is to train and construct embeddings by running encoders directly on individual documents. In this work, we argue that these embeddings, while effective, are implicitly out-of-context for targeted use cases of retrieval, and that a document embedding should take into account both the document and neighboring documents in context – analogous to contextualized word embeddings. We propose two complementary methods for contextualized document embeddings: first, an alternative contrastive learning objective that explicitly incorporates document neighbors into the intra-batch contextual loss; second, a new contextual architecture that explicitly encodes neighbor document information into the encoded representation. Results show that both methods achieve better performance than biencoders in several settings, with differences especially pronounced out-of-domain. We achieve state-of-the-art results on the MTEB benchmark with no hard negative mining, score distillation, dataset-specific instructions, intra-GPU example-sharing, or extremely large batch sizes.  Our method can be applied to improve performance on any contrastive learning dataset and any biencoder.


\end{abstract}

\section{Introduction}


Machine learning approaches to text retrieval aim to learn an embedded representation for indexing documents. Classically, this area was dominated by statistical approaches using sparse lexical matching methods based on n-gram frequencies such as BM25~\citep{robertson2009probabilistic}. Only recently have neural networks become competitive with state-of-the-art models on retrieval tasks~\citep{karpukhin2020dpr,thakur2021beir}. The primary neural method is a \textit{dual encoder} architecture that independently encodes both a document and query to a dense latent space for retrieval lookup. This document embedding space can improve upon a statistical model since it is learned end-to-end for retrieval.

However, there is at least one notable benefit of statistical approaches that is lost by neural models.  Statistical models can easily incorporate prior corpus statistics such as inverse document frequency (IDF), into their representation. This prior term imparts context-dependence onto the model, since it can be updated  based on information specific to retrieval in a given domain at test time. We contrast this contextual formulation with neural document encoders that are by definition a function of the document itself. For example consider the following document: 

\begin{quote}    
\label{quote:runex}
 The National Football League Draft is an annual event in which the National Football League (NFL) teams select eligible college football players...
\end{quote}

Depending on the retrieval domain, e.g. Wikipedia search, sports articles, or televised events, IDF may weight terms such as \texttt{NFL}, \texttt{draft} or \texttt{annual} higher; a neural document embedding model would need to select a global weighting for this document. 

In this work, we explore contextualization of document embeddings produced by dense encoders. The goal is to produce embeddings that are better able to handle retrieval tasks in specific challenging contexts. We propose two complementary changes to document encoders: a contextual training procedure and architecture.


For contextual training, we aim to build a notion of neighboring documents directly into the contrastive learning process. We propose a method that uses fast query-document clustering to produce a group of neighbors for each training batch. Each update for training is constructed purely from neighboring documents to ensure that embeddings can distinguish documents even in the most challenging contexts. 

\begin{figure}[t]
    \includegraphics[width=\textwidth]{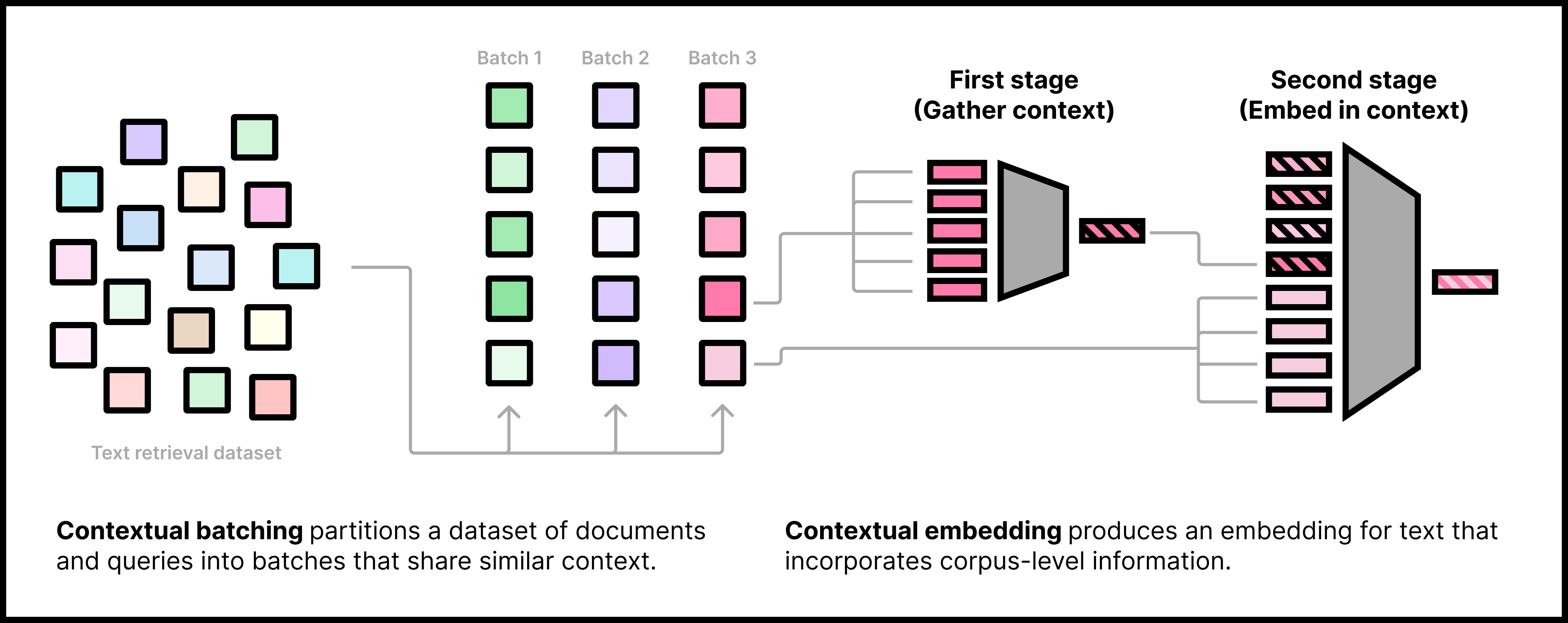}
    \label{fig:main}
    \caption{Overview of our system for contextual document embeddings (CDE). Our model operates in two stages: a first stage used to characterize the dataset from samples, and a second stage used to embed the final document.}   
\end{figure}

For the architecture, we propose a new encoder that injects information about the contextual documents during embedding. The proposed architecture augments the standard BERT-style encoder with additional conditioning that provides aggregated document-level information about neighboring documents. We call our method \underline{C}ontextual \underline{D}ocument \underline{E}mbedding (CDE). Analogously to pre-computed corpus-level statistics, this method provides a manner for the embedding to take into account the relative frequency of terms in context. The final output is still an embedding of the same size, so this does not require any additional storage or other changes to the retrieval process. When indexing, we utilize information from the corpus to produce document and query embeddings that are specific to a particular domain.

Experiments compare these two extensions to standard approaches for training document embeddings. Our results show that contextual contrastive training improves standard text embedding model training, and can be run without other approaches such as additional hard negatives. With the contextual encoder architecture, we see additional improvements over a baseline model in all settings tested, with larger improvements in highly specific domains such as small datasets of financial and medical documents. When trained at industry-scale, our model achieves state-of-the-art results for small ($<$250M parameter) models on the MTEB benchmark.



\section{Related Work}

\paragraph{Text retrieval.} Our work is related to the general field of text retrieval; we propose specific improvements to the training of ``biencoder'' text embedding models such as DPR \citep{karpukhin2020dpr}, GTR \citep{ni2021gtr}, Contriever \citep{izacard2022unsupervised}, LaPraDoR \citep{xu2022laprador}, Instructor \citep{su2023embedder}, Nomic-Embed \citep{nussbaum2024nomic}, E5 \citep{wang2024e5}, and GTE \citep{li2023gte}.
We focus on the problem of adapting these text retrieval models to new corpora at test time; some prior work has noted this problem \citep{dai2022promptagator, sciavolino2021universal} and proposed solutions such as unsupervised span-sampling and training on test corpora \citep{gao2021unsupervised} and distillation on the test corpus from a reranker \citep{sung2023optimizing}. Late interaction methods \citep{khattab2020colbert, santhanam2022colbertv2} also offer one way to improve out-of-domain retrieval performance, but increase the runtime and complexity of search. We propose a better sampling scheme that can be used to train any biencoder or late interaction model as well as a \textit{training-free} method for test-time adaptation.

\paragraph{Contrastive learning.} Much research has focused on the effect of hard negatives on the performance of contrastive learning methods \citet{chen2020simclr, qu2021rocketqa, robinson2021contrastivehn, wang2023simlm}. \citep{zhang2021understandingnce} observe that harder negatives provide a better approximation of the overall cross-entropy loss, but do not consider \textit{batch}-level optimizations for negative selection. \citet{hofstätter2021efficientlytasb} cluster queries before training and show that this improves performance. \citet{sachidananda2023globalselection} also consider contrastive batch sampling as a global optimization problem, but do not apply their technique to state-of-the-art transformer-based text embedding models.  \citep{ma2024mode} use a clustering algorithm to partition a dataset into several sub-datasets, but train a different model on each sub-dataset. \citet{solatorio2024gistembedguidedinsampleselection} also use a pre-trained model to address the problem of in-batch false negatives from randomly sampled batches. Our training algorithm aims to find the hardest possible high-quality batches to train text embedding models.

\paragraph{Test-time adaptation.} Our method can be compared to other solutions to test-time adaptation, a problem that has been well-studied across a variety of domains \citep{jang2023testtimeadaptation}. In retrieval, one form of test-time adaptation is pseudo-relevance feedback (PRF) \citep{rocchio1971prf, li2018nprf, wang2021colbertprf}, where documents relevant to the query are used to construct a final, enhanced query representation. The query side of our model can be seen as a form of pseudo-relevance feedback; however, we train from scratch to support a more general form of PRF natively, on the document representation as well as the query.

\paragraph{Non-parametric modeling.} Our contextual document model can be seen as a form of non-parametric modeling. This shows connections with the a large body of deep learning research such as the non-parametric transformer (NPT) \citep{kossen2022selfattention} and the subfield of Neural Processes \citep{garnelo2018conditional, kim2019attentive, nguyen2023transformernp}.
Semi-parametric models have been recently applied in NLP, specifically to the task of language modeling \citep{borgeaud2022retro,khandelwal2020knnlm}. Instead of using a retrieval model to build a semi-parametric langauge model, we build a semi-parametric model specifically for the task of retrieval.


\section{Background}

We can view text retrieval methods probabilistically as computing a distribution over potential documents based on a scalar score function $f(d, q)$ matching documents and queries:
\begin{equation}
    p(d \mid q) = \frac{\exp{f(d, q})}{\sum_{d' \in {\cal D}} \exp{f(d', q})}
    \label{eq:main}
\end{equation}
where ${\cal D}$ is a finite set of documents in a dataset. There is a wide variety of different definitions for $f$ including full pairwise neural parameterizations~\citep{nogueira2020passage}. In this work, we focus on efficient retrieval methods using vector-based methods, also known as embedding models. 

Vector retrieval methods assume that $f(d, q)$ can be factored into two embedding terms, $\phi(d) \cdot \psi(q)$, the document and query embedding respectively. This factorization allows precomputation of the document embeddings $\phi(d)$ for all $d \in {\cal D}$. This is critical for facilitating fast computation of $\arg\max_{d} p(d \mid q)$ or top-k variants~\citep{douze2024faiss}. 

In statistical retrieval, $\phi$ and $\psi$ are closed-form functions of the data, often representing unigram or bigram counts by the relative frequency of word types. Notably for this work, these methods can also utilize distributional properties of the test dataset as a prior, for example through inverse document frequency (IDF). We represent this integration of dataset-level information by writing the vector product $\phi(d;{\cal D}) \cdot \psi(q; {\cal D})$.

In neural retrieval, we instead learn the representation as a dense vector. We assume access to a training corpus of document and query pairs (these may be supervised, i.e. gold-standard annotations, or unsupervised, i.e. noised synthetic examples), ${\cal D}_T = \{(d^1, q^1), ..., (d^J, q^J)\}$, with the aim of learning the embedding function $\phi$ and $\psi$. 

Training can be motivated as maximizing likelihood of the document corresponding to each query, i.e.  $\sum_j\log p(d^j \mid q^j)$. Unfortunately, since retrieval datasets can have $|{\cal D}|$ exceed millions of documents, computing the normalizer in Eq~\ref{eq:main} at each training step is not an option. Instead contrastive learning is used where the likelihood is replaced with a biased approximation calculated from negative samples: 

\[ \max_{\phi, \psi} \sum_j \log p(d^j \mid q^j) \approx \sum_j \log \frac{\exp{f(d^j, q^j})}{\sum_{d' \in {\cal H}(q^j)} \exp{f(d', q^j})} \]

where ${\cal H}$ is a set of examples used to approximate the normalizing constant. In implementation, in addition to these hard negative examples, other examples from the mini-batch are also used to compute the normalizer since it requires no additional compute for calculating $\phi(d)$.

\section{Methods}

In our work, we are interested in integrating contextual information into our embedding functions $\phi$ and $\psi$. The standard neural $\phi$ is purely a function of the document $\phi(d)$ and does not take into account any notion of context. This contrasts with the statistical model $\phi(\cdot; {\cal D})$ and $\psi(\cdot; {\cal D})$. Arguably this is not an issue if retrieval is completely in domain, as $\phi$ is capable of learning statistics such as IDF and average document length on the training set through gradient descent.

However, in many retrieval benchmarks, models are trained over a single set of documents $\cal D$ and then tested in many other domains ${\cal D}$ that differs significantly from ${\cal D}_T$ . In this setting, training on ${\cal D}_T$ alone may not be able to provide robust embeddings when used in contexts such as ${\cal D}$.

\subsection{Contextual Training with Adversarial Contrastive Learning}

Returning to the example from the introduction, we assume that in a general purpose training corpus ${\cal D}_T$, the term \texttt{NFL} is a rare word appearing in relatively few documents and a useful signal. However, if at test time ${\cal D}$ is a corpus of sports articles, this word would be exceedingly common. Evaluation in this domain is, in a statistical sense, adversarial to the original dataset. To handle this issue, meta-learning-style objectives have shown to be effective for training document embedders. In these approaches, instead of sampling documents-query pairs iid, the objective first sample a domain and then sample a batch of examples. This ensures that the model mostly sees related training points in each domain.

We propose a training objective that synthesizes a large set of fine-grained domains to train the model on.  Formally, our aim is to partition the training dataset ${\cal D}_T$ into groups $({\cal B}^1, \ldots {\cal B}^B)$ such that each group represents a self-similar pseudo-domain:

\[ \max_{\phi, \psi} \sum_{b }\sum_{(d, q) \in {\cal B}^b } \log p(d \mid q) =  \max_{\phi, \psi} \sum_b \sum_{(d, q) \in {\cal B}^b} \log \frac{\exp{f(d, q})}{\sum_{(d', \cdot) \in {\cal B}^b} \exp{f(d', q})} \]

Computationally, the inner term can be implemented as a single batch and computed efficiently without the need for separate hard negatives ($\mathcal{H}$).
Ideally we want groups that are as challenging as possible.
\citet{zhang2021understandingnce} show that increasing the partition term improves the contrastive approximation to the maximum likelihood of the gradient. We can formalize this search for the most difficult configuration of batches as an optimization problem:

\begin{equation}
\max_{({\cal B}^1, \ldots {\cal B}^B)} \sum_b \sum_{\substack{(d, q) \in {\cal B}^b \\ (d', q') \in {\cal B}^b }} \left( f(d, q') + f(d', q) \right) =  \max_{({\cal B}^1, \ldots {\cal B}^B)}  \sum_b \sum_{\substack{(d, q) \in {\cal B}^b \\ (d', q') \in {\cal B}^b }} \left( \phi(d) \cdot \psi(q')  + \phi(d') \cdot \psi(q)  \right)
\end{equation}

Solving this combinatorial objective exactly is intractable, but we can approximate a solution using clustering. We first move from a maximization to a minimization by replacing the two dot products with L$_2$ distance $m((d,q), (d', q')) = ||\phi(d) - \psi(q')|| + ||\phi('d) - \psi(q)||$ (which is equivalent for normalized embeddings). We then note when that treated as symmetric pairs, this term obeys the triangle inequality for any other pair $m$: 

\[ m((d,q), m) +  m(m, (d',q')) \geq m((d,q),(d', q'))  \] 

This implies that the following centroid-based objective represents an upper-bound on our original objective:
\begin{equation}
\min_{\substack{({\cal B}^1, \ldots {\cal B}^B) \\ (m^1, \ldots, m^B)}} \sum_b \sum_{\substack{(d, q) \in {\cal B}^b}} m((d,q), m^b)
\end{equation}
For known $B$, this search defines an asymmetric K-Means clustering problem. A solution can be efficiently computed using extremely fast  Euclidean K-Means packages be treating each data point as two separate vectors $\phi(d) \oplus \psi(q)$ and $\psi(q) \oplus \phi(d)$, where $\oplus$ is concatenation.

\paragraph{Cluster Embeddings.}
 Since clustering is performed before training, we do not have dense encoders $\phi$ and $\psi$ when constructing the groups. Borrowing methods from hard-negative mining~\citep{robinson2021contrastivehn} we can replace the $\phi$ and $\psi$ with a simpler embedding model when constructing groups. We experiment with a sparse vector representation and with pretrained dense representations, settling on GTR \citep{ni2021gtr}, a popular and generic text embedding model.

\paragraph{Filtering False Negatives.} 

Our method is especially sensitive to false negatives, as they will be more likely to be included in a given batch.  Unfortunately, traditional retrieval datasets are not designed with this type of global objective in mind: false negatives are common in most retrieval datasets and their prevalence increases with dataset scale. As one datapoint, \citet{qu2021rocketqa} found that over 70\% of top-retrieved passages in MS Marco are false negatives.

To avoid a situation where each batch contains a large number of false negatives, we compute an equivalence class: $S(q, d) = \{ d' \in \mathcal{D} \mid f(q, d') \geq f(q, d) + \eps \}$ for some surrogate scoring function $f$ and boundary term $\eps$. At training time,  we alter the partition function for $d$ so that it no longer includes the elements of $S(q, d)$, which are not definitively negative examples:
\begin{equation}
    \log p(d \mid q) = \dfrac{\exp{f(d, q)}}{\exp{f(d, q)} + \sum_{d' \notin S(q, d)} \exp f(d', q)}
\end{equation}
For simplicity, we again select $f$ to be a simple pre-trained embedding model. This method likely over-prunes some potential true negatives found by the surrogate model; however we found it to be critical to model accuracy. 

\paragraph{Packing.} Clusters found by our algorithm will be of varying sizes, and need to be packed into equal-sized batches. We apply a post-hoc procedure. We consider both random partitioning and grouping via greedy cluster-level traveling salesman, similar to \citet{shi2024incontextpretraining}. In both cases, we split large group into into smaller batches, and merge close small batches from within the same domain into evenly-sized batches. This has an added benefit of introducing randomness into the groups when training for multiple epochs. We leave it to future work to analyze the full effects of different packing strategies such as expensive Balanced K-Means  or heuristic approaches such as Equal K-Means~\citep{gururangan2023scalinglm}.

\subsection{Contextual Document Embedding (CDE)}

Contextualization can also be added directly to the architecture. Taking inspiration from sparse vector retrieval which uses corpus statistics to determine the form of the embedding, we modify the encoders to have access to the corpus itself, i.e. $\phi(d; {\cal D})$ and $\psi(d; {\cal D})$. This effectively 
 augments  the biencoder model to give it the ability to contextualize documents directly.

The main challenge is how to design a neural architecture that can take into account dataset contextualization. On one extreme, we could follow methods like BM25 and precompute a fixed set of corpus statistics that could be fed to the document encoder. On the other extreme, we could allow the encoder full access to the entire corpus, through some form of cross attention. The latter approach has been explored on a small scale in methods like neural processes~\citep{garnelo2018conditional}; however, it would be difficult to scale to larger datasets.  

We opt for a middleground that allows the model to learn corpus statistics, but is also relatively efficient to compute, shown in Figure~\ref{fig:main}. Specifically, we note that document embeddings retain a surprising amount of lexical information even after embedding \citep{morris2023teraamat}. Therefore, if we pre-embed a subset of the corpus, we believe we can still dynamically calculate key dataset information during encoding.

We produce contextualized embeddings via a two-stage process:

\textit{\textbf{First stage:} Gather and embed context.} Given context documents $d^1, ..., d^J \in \mathcal{D}$, we embed each using a unique embedding model and concatenate embeddings into a sequence $M_1(d^1) ... M_1(d^J)$. 

\textit{\textbf{Second stage:} Embed document with additional context tokens.} To compute $\phi$ for document $d'$ we integrate contextual embedding sequence at the input of second-stage embedding model $M_2$:
\begin{equation}
    \phi(d'; \mathcal{D}) = M_2(M_1(d^1), \ldots, M_1(d^J), E(d'_1), \ldots, E(d'_T))
\end{equation}

Here $M_1$ is the first-stage encoder model, $M_2$ is a second-stage encoder model, and $E$ is the token embedding matrix of $M_2$ applied to each token in $d'$. In practice, we parameterize both $M_1$ and $M_2$ using traditional bidirectional transformers, so our model is comprised of two biencoder-like backbones called in sequence. 

There is a similar contextualized model for the query encoder $\psi$ which is also given document context (as we do not have query context at test time):
\begin{equation}
    \phi(q; \mathcal{D}) = M_2(M_1(d^1), \ldots, M_1(d^J), E(q_1), \ldots, E(q_T))
\end{equation}

We note several implementation properties of this architecture. During training, computing contextual embeddings for each contextual document for each training instance would naively increase training by a computational factor proportional to $J$, the number of documents in context. This time increase would not be tractable, since contrastive training can already take many days. We overcome this difficulty by sharing context $d^1, ..., d^J$ within a batch of documents; this allows us to compute representations just once per training step and reuse them between documents via computational graph. \footnote{Context reuse is only feasible because documents within the same batch typically share a large amount of context anyway, since they are clustered.}

When indexing a new corpus $\cal D$, first stage representations $M_1(d^1) ... M_1(d^J)$ can be computed once and cached, so $M_1$ does not add parameters or runtime to the search process. Query representations can also use the cached context, which only require additional inputs to the encoder. (Our model does not include contextualized queries, only documents, as we typically do not assume access to example queries at test-time.)

\paragraph{Embedding \textit{without} context.} Individual corpora during training may not have sufficient or available context. To improve our model's generalization, we use \textit{sequence dropout}, where we randomly replace context embeddings $M_1(d^*)$ with some null token $v_\emptyset$ according to some a uniform probability $p$.

At test time, if no corpus information is available, our model can now function as a non-contextual biencoder simply by replacing all sequence token inputs with $v_\emptyset$.

\paragraph{Position-agnostic embedding.} Since documents of $\mathcal{D}$ are unordered, we remove all positionality from the neural encodings. When parameterizing $\theta$ with a traditional transformer, this can be achieved by omitting positional embeddings at the positions corresponding to $\mathcal{D}$. In practice, we use transformers implementations dependent on FlashAttention with rotary positional embeddings at each self-attention layer. Full details of how we disable positionality are available in \Cref{app:flash-attention-positionity-disabling}.

\paragraph{Two-stage gradient caching.} To improve training we employ a gradient-caching technique analogous to a two-stage version of GradCache \citep{gao2021gradcache}. This technique allows us to fit larger batches, longer sequences with more contextual samples without running out of memory. Essentially, we compute first-stage and second-stage representations independently without gradients. We then use these frozen representations to compute the loss, and gradients with respect to the second-stage representations. We then re-run the second stage with gradients enabled and use the output gradients to backpropagate through the second-stage model, and obtain gradients for the first-stage representations. We repeat this process for the first-stage representations. This allows us to tradeoff computation (running each transformer forward pass twice) for memory. 

\section{Experimental Setup}

We consider a range of retrieval experiments across different scales. To run experiments across a suitable number of settings, we devise a small setting: six-layer transformer, maximum sequence length of $64$, and maximum number of $64$ additional contextual tokens. In this scenario, we evaluate on a truncated version of the BEIR benchmark \citep{thakur2021beir}. Given the low cost of each experiment, we are able to pre-train and fine-tune both biencoder and contextual models across a variety of batch sizes in $\{256, 512, 1024, 2048, 4096\}$ and cluster sizes $\{64, 256, 1024, 4096, ..., 2097152, 4194304\}$. As typical state-of-the-art text embedding models are trained in two phases, a large weakly-supervised pre-training phase and a short supervised phase, we run all experiments for both phases. 

For the large setting, we use the best settings found via small experiments. We train a single model on sequences of length $512$ with $512$ contextual documents, evaluating on the full MTEB benchmark \citep{muennighoff2022mteb}. This includes tasks from retrieval as well as tasks like classification, clustering, and reranking. We compare our model's performance to the top small-size (under 250M parameters) models on MTEB \citep{nussbaum2024nomic, xiao2024bge, solatorio2024gistembedguidedinsampleselection, li2023gte}.

\paragraph{Training Data and Metrics}
We train on the meta-datasets collected in \citet{nussbaum2024nomic} for training text embedding models. This collection of datasets includes data from $24$ datasets scraped from web sources such as Wikipedia and Reddit. Our unsupervised training phase trains on 200M weakly-supervised datapoints scraped from large internet sources such as Reddit and Wikipedia. The supervised training phase includes 1.8M human-written query-document pairs intended for text retrieval, and is aggregated from popular retrieval datasets such as HotpotQA and MS MARCO \citep{yang2018hotpotqa, bajaj2018msmarco}. For our full model, we also consider supervised training on the BGE meta-datasets \citep{xiao2024bge}. We evaluate our models using NDCG@10, a conventional retrieval metric that enables comparison across many disparate datasets.

\paragraph{Implementation}
When partitioning our dataset into batches, we encode documents and queries using GTR \citep{ni2021gtr} and implement our clustering algorithm on top of FAISS \citep{douze2024faiss}. We cluster per-domain for $100$ steps and take the best clustering out of $3$ attempts.
We select NomicBERT as our pre-trained model backbone \citep{nussbaum2024nomic}, which has 137M parameters. We prepend all texts with short task-specific prefixes to identify each task; prefixes are listed in \Cref{app:task-prefixes}. When pooling, we pool over text tokens only, never contextual tokens.

\paragraph{Training} We initialize both $M_1$ and $M_2$ using the BERT-base model from \citet{nussbaum2024nomic} that includes flash attention. Weights are shared between $\phi$ and $\psi$, but notably not between $M_1$ and $M_2$. For all experiments, we train with the Adam optimizer with 1000 steps of warmup to a learning rate of $2 \cdot 10^{-5}$ and linearly decay 
to $0$ throughout training. For the filtering model we select \texttt{nomic-embed-v1} which was trained on the same datasets \citep{nussbaum2024nomic}. We train for three epochs unless otherwise specified. We set the maximum sequence length for all inputs to $512$ and the number of contextual inputs to $512$ (so the second-stage model has an input length of $1024$). When computing contrastive loss, we use a fixed temperature of $\tau = 0.02$. When sequence dropout is enabled in our contextual architecture, we set contextual input tokens to null vectors with a uniform probability $p=0.005$. If the batch size exceeds the number of contextual documents, we randomly sample to produce contextual inputs. 

\section{Results}

\begin{table}[]
    \centering

\begin{tabular}{cc|cc|cc|c}
    \toprule
     \multicolumn{2}{c}{Contextual} \\
     Batch& Arch & Batch Size & Cluster Size & Train loss & Train acc. & NDCG@10 \\ \midrule
    & & 16384 & -&  0.39 & 90.3 & 59.9 \\
     \ding{51} & & 512 & 512 & 0.81 & 77.7 & 61.7\\
    &\ding{51} & 16384 & -& 0.37 & 90.7 & 62.4 \\
     \ding{51} & \ding{51} & 512 & 512 & 0.68 & 80.9   & \textbf{63.1} \\ \bottomrule
\end{tabular}

\caption{Performance of our small models with and without the two improvements proposed in this paper, measured on a shortened version of the BEIR benchmark. Numbers are NDCG@10.}
\label{tab:toy-results-beir}
\end{table}

The main results are highlighted in \Cref{tab:toy-results-beir} and \Cref{tab:mteb-full}. In the smaller setting, we observe that both adversarial contrastive learning and our contextual architecture improve performance compared to vanilla biencoder training. We observe the largest improvement when we combine these techniques. 

\begin{figure}[t]
    \centering
    \begin{minipage}{\textwidth}
        \begin{minipage}{\textwidth}  
            \includegraphics[width=\textwidth]{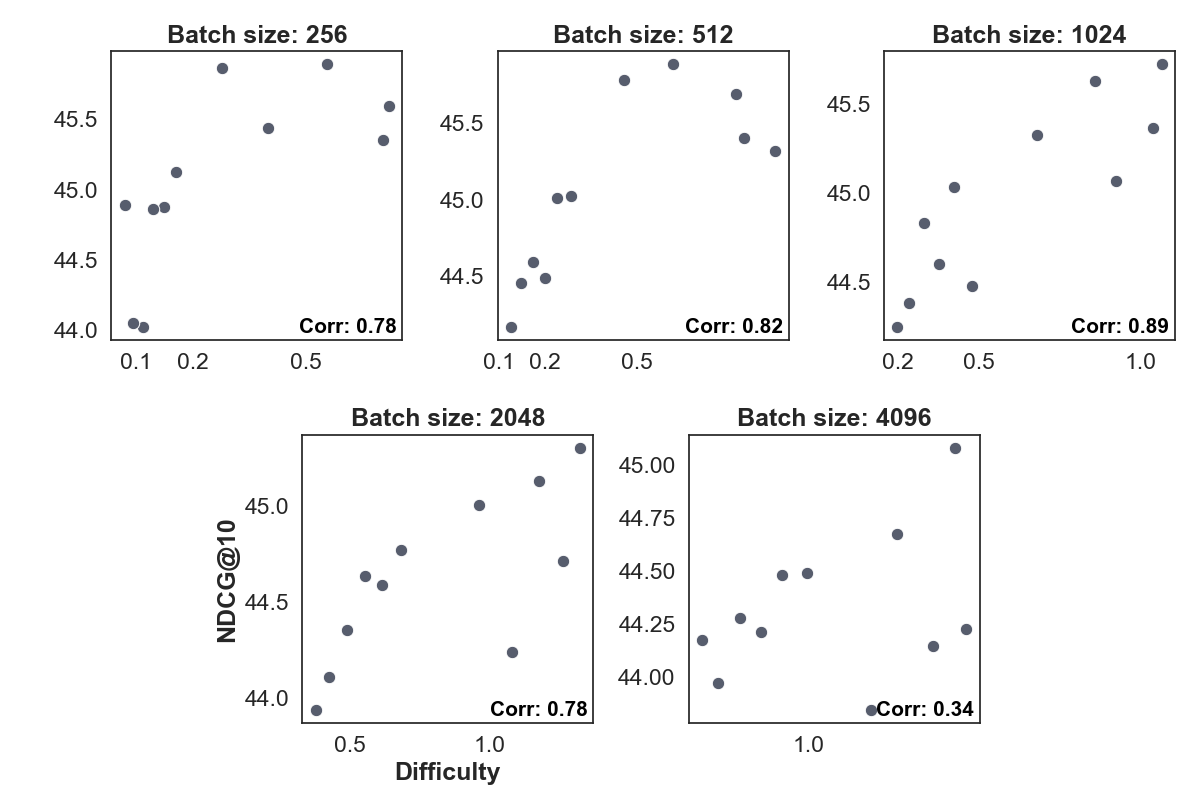}
            \caption{Performance vs. average batch difficulty (as measured by loss at the end of pre-training and supervised training) across batch sizes, after supervised contrastive training. Within a given batch size, we observe a clear increase in performance by making individual batches harder. Correlations are Pearson.}
            \label{fig:result_hardness_correlations}
            \vspace{.5cm}
        \end{minipage}
        \begin{minipage}{\textwidth}
            \begin{minipage}{\textwidth}
                \includegraphics[width=.48\textwidth]{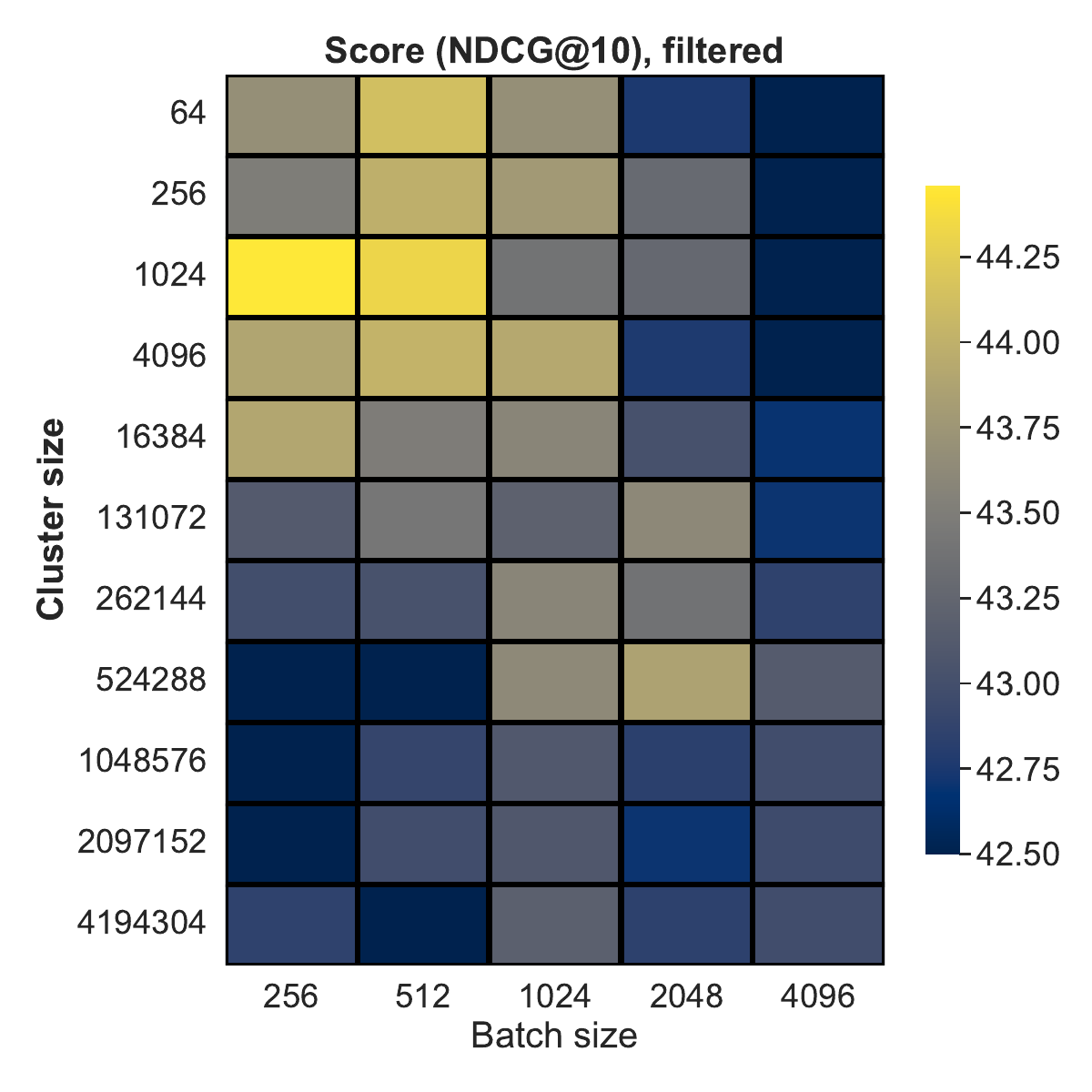}
                \includegraphics[width=.48\textwidth]{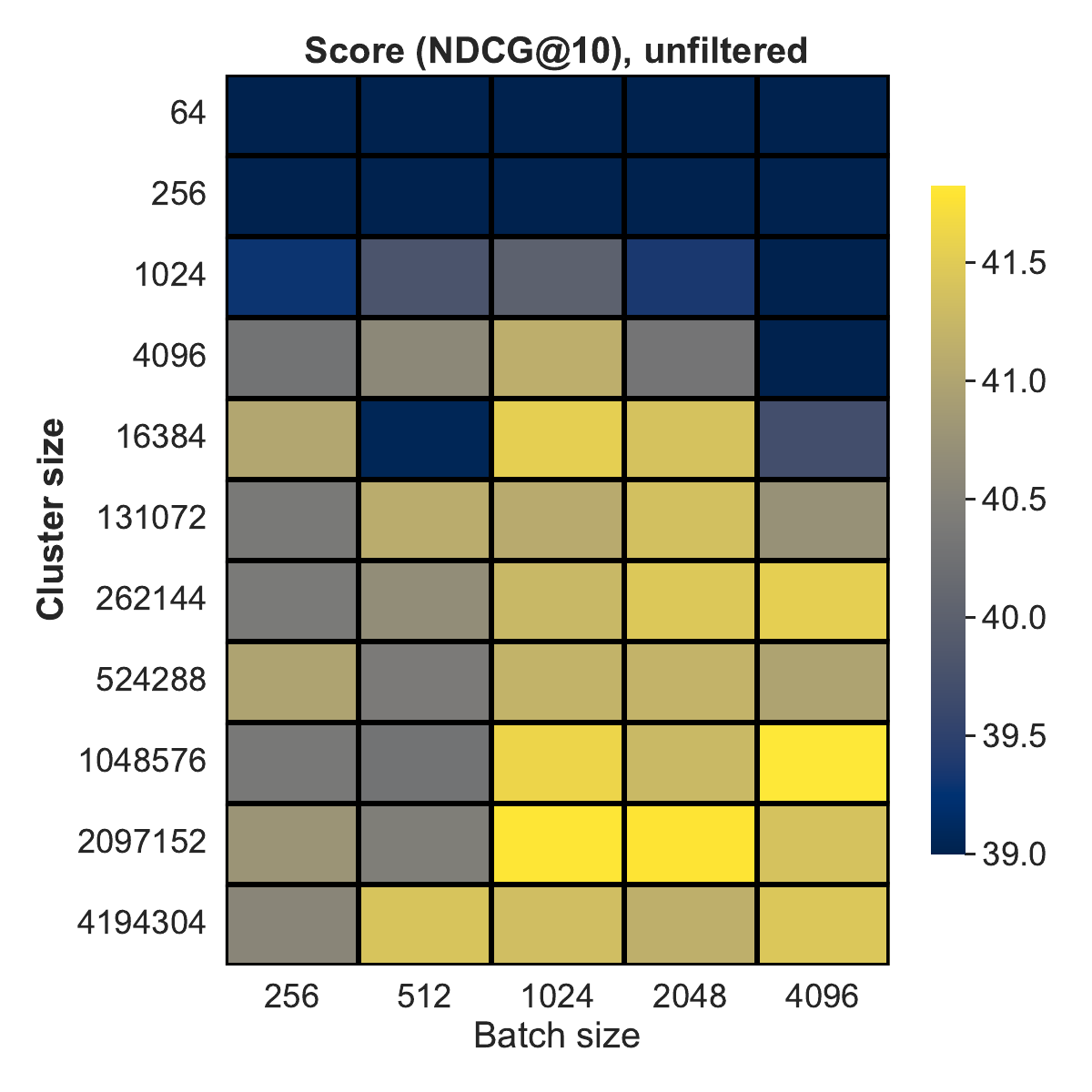}
            \end{minipage}
            \label{fig:result_filtering_heatmaps}
            \caption{Biencoder performance with filtering (left) and without (right) across batch and cluster sizes during unsupervised contrastive pre-training. With filtering, small cluster sizes clearly improve performance, and larger batch sizes do not.}
        \end{minipage}
    \end{minipage}
\end{figure}

\paragraph{Contextual batching} After controlling for batch size and filtering for false negatives, we observe a strong correlation (visualized in \Cref{fig:result_hardness_correlations}) between batch difficulty and downstream performance: \textit{reordering datapoints to make batches harder definitively enhances overall learning}. This corroborates prior findings \citep{xiong2020ance, qu2021rocketqa} and theory \citep{zhang2021understandingnce} that more difficult batches in contrastive learning form a better overall gradient approximation and learn more effectively. 

\begin{figure}[ht]
    \centering
    \hfill
    \begin{minipage}{0.48\textwidth}
        \includegraphics[width=\textwidth]{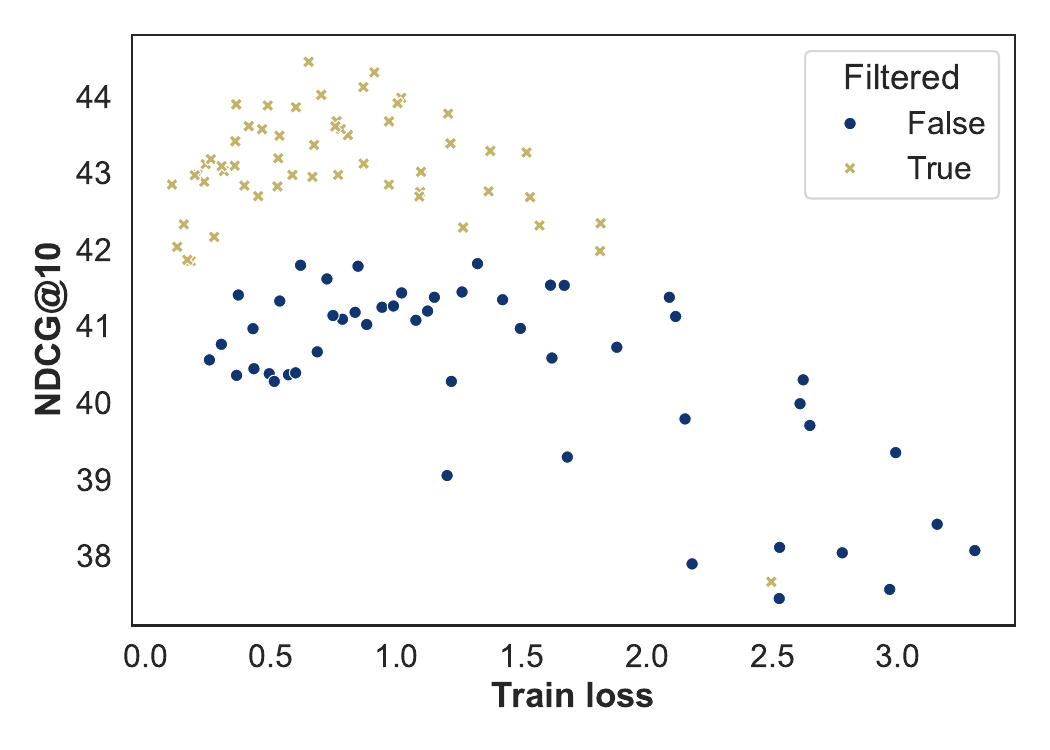}
        \caption{Impact of filtering during training across various batch and cluster sizes. Each dot is a biencoder pretrained with a different batch and cluster size.}   
        \label{fig:result_filtering_scatter}
    \end{minipage}
    \hfill
    \begin{minipage}{0.48\textwidth}
        \includegraphics[width=\textwidth]{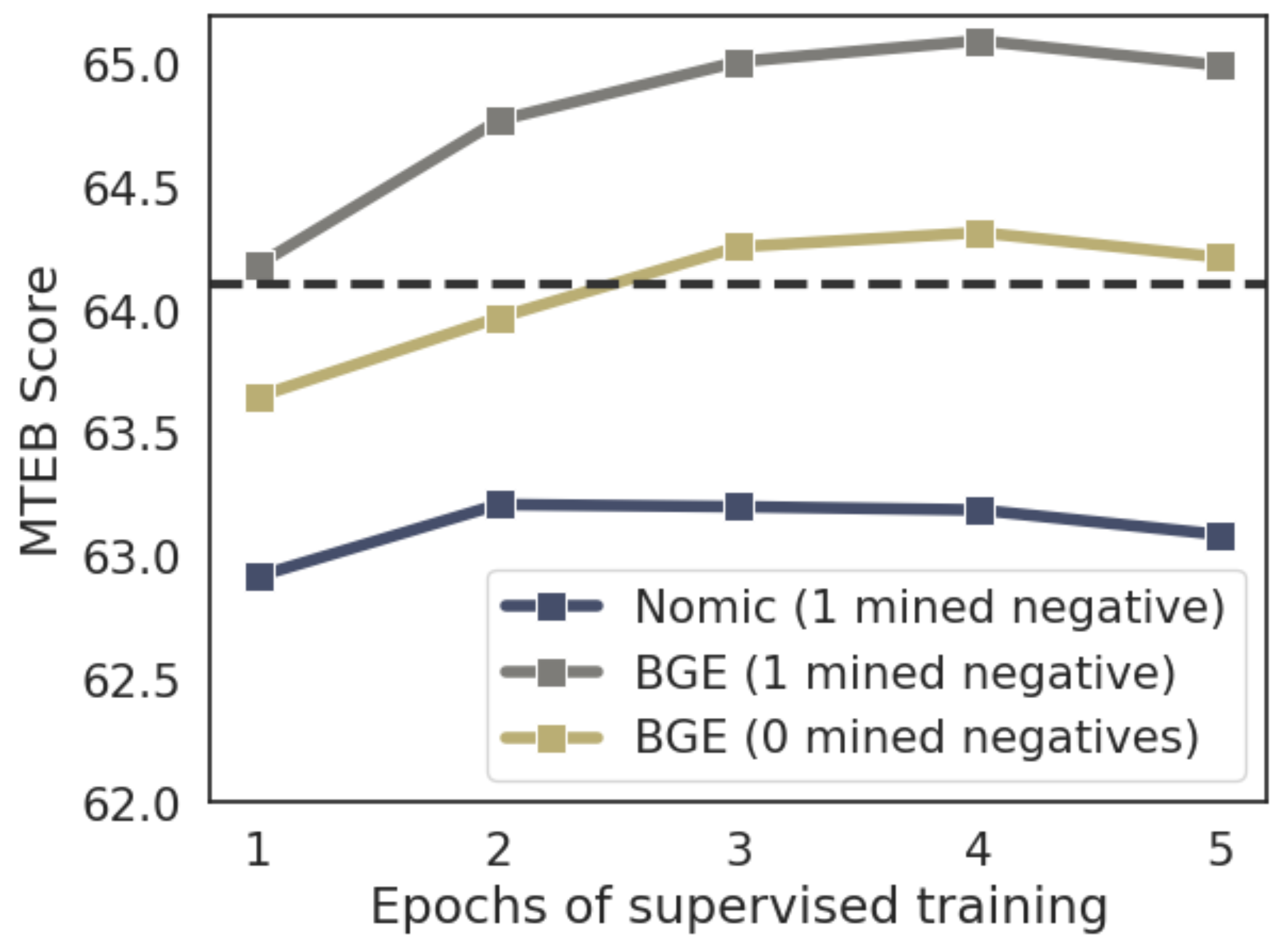}
        \caption{Performance on MTEB across epochs of supervised training on the Nomic and BGE supervised meta-datasets.}
        \label{fig:supervised_epoch}
    \end{minipage}
    \hfill
\end{figure}

\Cref{fig:result_filtering_heatmaps} showcases model performance across batch and cluster sizes after both phases of training. We observe that although a large batch and cluster size are useful when filtering is not enacted, when including filtering, smaller cluster (and harder) are clearly better, and large batches do not add much. When comparing filtered to non-filtered models (\Cref{fig:result_filtering_scatter}), filtering false negatives clearly improves performance.

\paragraph{Contextual architecture} In addition to adversarial batching, we compare our contextual architecture to a biencoder across the datasets of BEIR in \Cref{tab:toy-results-beir} (full results in appendix). Our architecture generally matches or improves performance on all downstream datasets, with largest improvements in ArguAna and SciFact, two of the smaller and more out-of-domain datasets.

\paragraph{Full-scale training}

\Cref{fig:supervised_epoch} shows our models' performance when trained for multiple epochs on the supervised datasets, relative to the best similar-sized embedding model (dashed line). We find best performance when training for four epochs on the BGE meta-datasets. Although our best model does use a single hard negative per query, we are still able to to achieve state-of-the-art performance without using \textit{any} hard negatives.

For our final model (\ModelName), we select the best of the supervised models, which comes from finetuning on the BGE dataset. On MTEB, \ModelName\ obtains state-of-the-art results compared to models of the same size. Although inspired by problems in the specific domain of text retrieval, we observe that our approach improves embedding performance in all domains, including clustering, classification, and semantic similarity. We also evaluate a ``random documents'' baseline, where we sample random documents from the training dataset to simulate a scenario where we lack access to the test corpus. In this setting, we drop around $1.2$ points on average across all tasks; the STS tasks in particular appear to produce representations that are close to context-agnostic.


\definecolor{maroon}{RGB}{0,0,128}  

\newlength{\colwidth}
\setlength{\colwidth}{.8cm}
\begin{table}[]
\centering
\begin{tabular}{p{3cm}p{\colwidth}p{\colwidth}p{\colwidth}p{\colwidth}p{\colwidth}p{\colwidth}p{\colwidth}p{\colwidth}@{}}
\toprule
 & Clssfctn & Cluster & PairCls & Rerank & Retrvl & STS  & Summ. & Mean  \\
\midrule
nomic-embed-v1 & 74.1 & 43.9 & 85.2 & 55.7 & 52.8 & 82.1 & 30.1  & 62.39 \\
stella-base-en-v2 & 75.3 & 44.9 & 86.5 & 58.8 & 50.1 & 83.0 & 32.5 & 62.61 \\
bge-base-en-v1.5 & 75.5 & 45.8 & 86.6 & 58.9 & 53.3 & 82.4 & 31.1 & 63.56 \\ 
GIST-Embedding-v0 & 76.0 & 46.2 & 86.3 & 59.4 & 52.3 & 83.5 & 30.9 & 63.71 \\
gte-base-en-v1.5 & 77.2 & 46.8 & 85.3 & 57.7 & 54.1 & 82.0 & 31.2 & 64.11 \\ 
\midrule
\ModelName{} \\
\ {\textcolor{maroon}{[Random]}} & 81.3 & 46.6 & 84.1 & 55.3 & 51.1 & 81.4 & 31.6 & 63.81 \\
\ {\textcolor{maroon}{[Contextual]}} & 81.7 & 48.3 & 84.7 & 56.7 & 53.3 & 81.6 & 31.2 & 
\textbf{65.00} \\
\bottomrule
\end{tabular}
\label{tab:mteb-full}
\caption{Performance of models with 250M or fewer parameters on the MTEB benchmark for text embedding models. ``Random'' indicates the performance of our model with random training documents included instead of per-domain contextual documents.}
\end{table}

\begin{figure}[ht]
    \begin{minipage}{0.48\textwidth}
        \includegraphics[width=\textwidth]{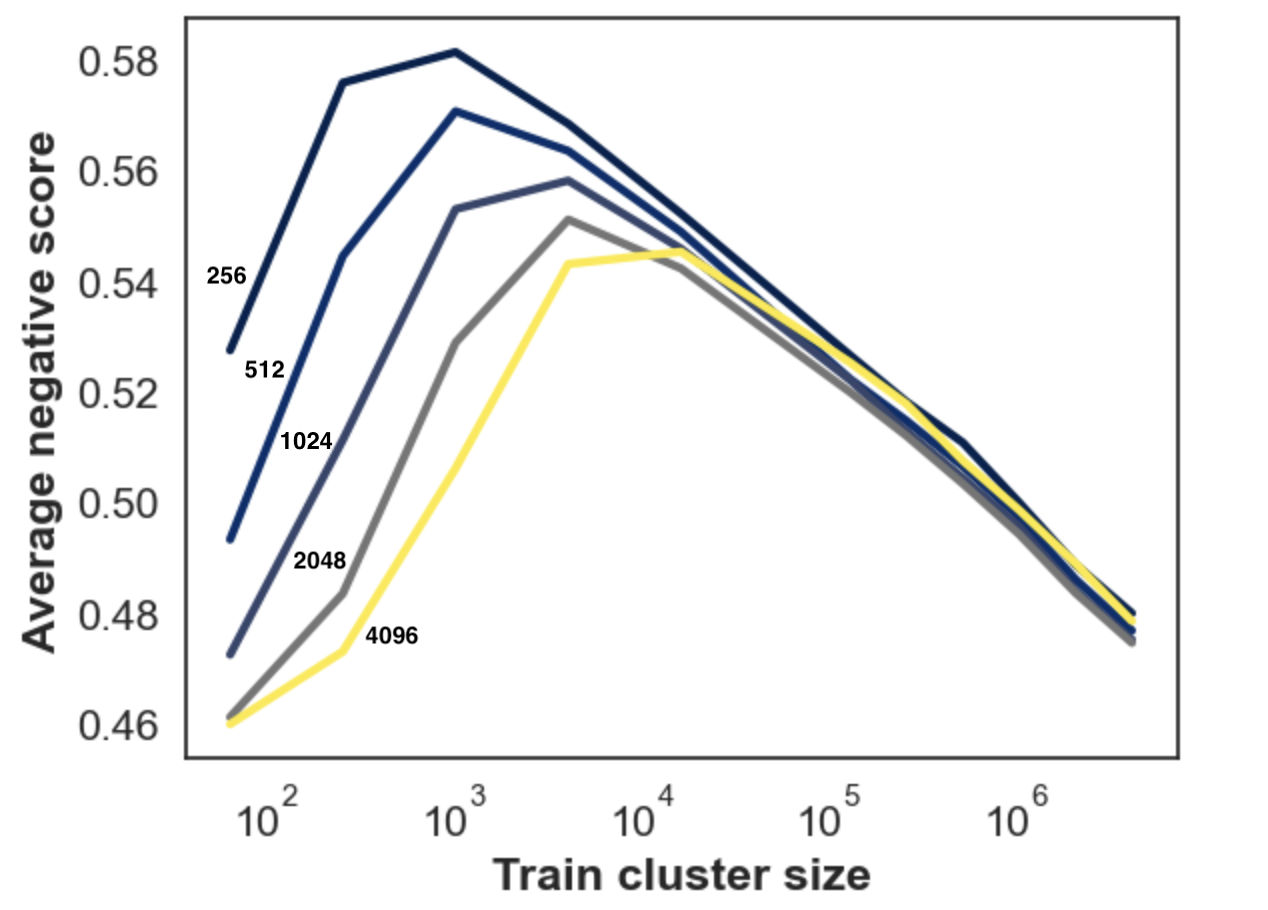}
        \caption{Average difficulty of in-batch negatives as measured by a surrogate model as cluster size and batch size change.}
        \label{fig:analysis_cluster_hardness}
    \end{minipage}
    \hfill
    \begin{minipage}{0.45\textwidth}
        \includegraphics[width=\textwidth]{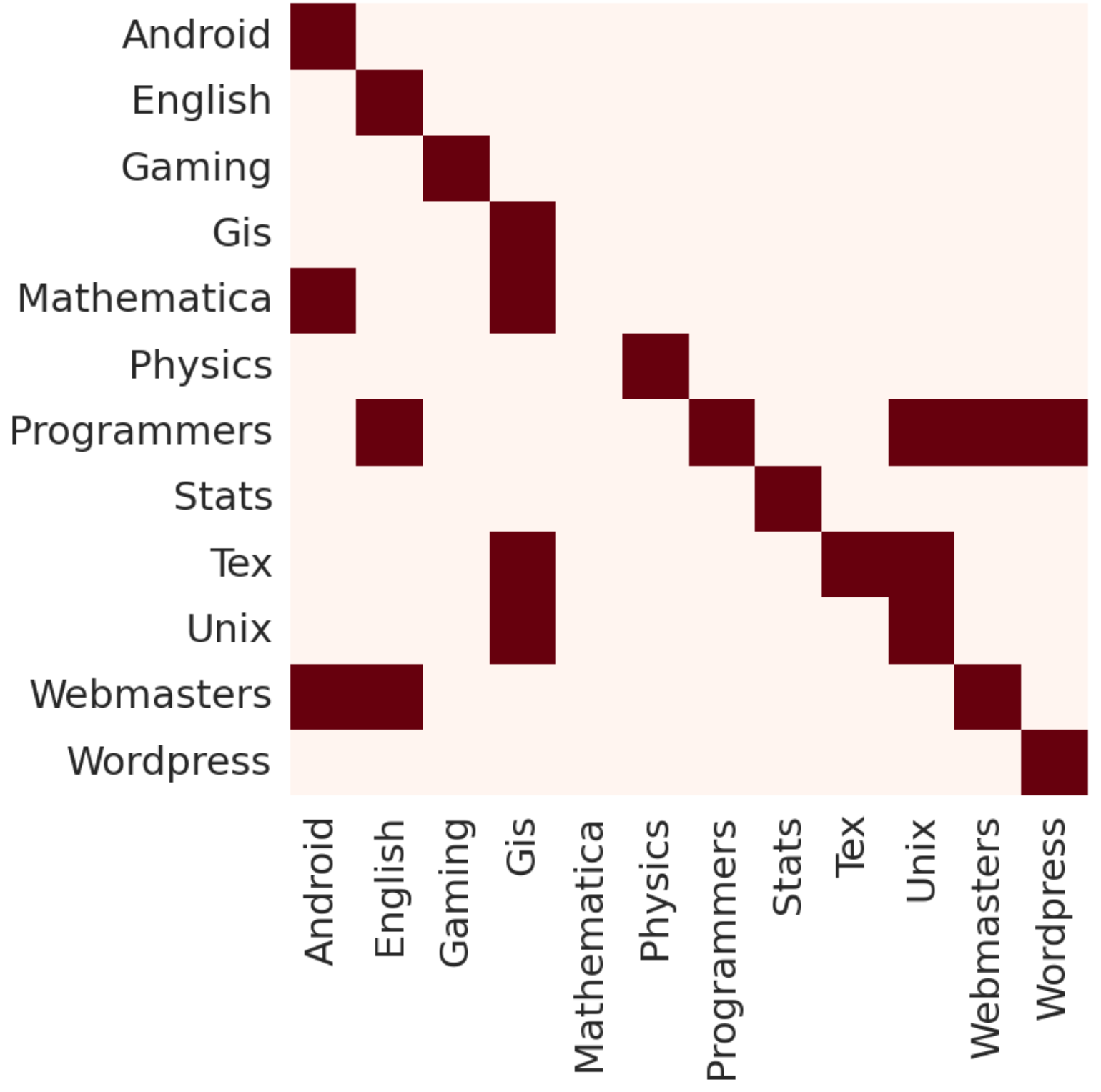}
        \caption{Impact of context by testing our model with different Stackexchange forum input types. Y-axis indicates the input domain, X-axis indicates the test domain. Dark squares come within one point NDCG@10.}
        \label{fig:cqadupstack}
    \end{minipage}
\end{figure}

\section{Analysis}

\paragraph{How hard are our clusters?}  To analysis the relationship between cluster size in our clustering algorithm and the overall average difficulty of in-batch negatives, we measure the average difficulty of 1000 batches across a variety of batch and cluster sizes and plot the data in \Cref{fig:analysis_cluster_hardness}. We observe that larger batches bring easier non-negative examples, and decreasing cluster size clearly increases the average hardness of negative examples in a given cluster.

\paragraph{Which contextual documents help?} To confirm that the CDE model is utilizing contextual information from $\cal D$ we consider how different contextual documents help for a given docuent $d$. \Cref{fig:cqadupstack} measures results on CQADupstack, a collection of Stack Exchange forum posts. We randomly sample inputs to from $\cal D$ from a domain (x-axis) and use them as input to the downstream task $d$ marked along the y-axis. We mark a square as red if its score comes within 1 point of NDCG of the top score for its domain. Generally utilizing in-domain works best, but there are some crossover interactions.

\section{Conclusion}
We propose two improvements to traditional biencoder models for generating embeddings. The first improvement involves an algorithm for reordering training datapoints to make batches harder and improves vanilla training with minimal changes. Our second improvement involves a new corpus-aware architecture for retrieval and allows us to train a state-of-the-art text embedding model.

\section{Acknowledgements}

Thanks to Orion Weller, Vin Sachidananda, and Zach Nussbaum for valuable feedback on this research. We would also like to acknowledge to Nomic and Hyperbolic for providing the compute necessary to conduct this research. This work was partially supported by Intelligence Advanced Research Projects Activity (IARPA), via the HIATUS Program \#2022-22072200003. JM is supported by an NSF GFRP fellowship.

\clearpage
\bibliography{iclr2025_conference}
\bibliographystyle{iclr2025_conference}

\clearpage
\section{Supplementary Material}

\subsection{Computational resource usage}
We pre-train all models on 8 NVIDIA H100 GPUs. In the slowest setting, training a biencoder for a single unsupervised epoch (235M pairs) takes approximately one day. Training our contextual archiecture for a single epoch takes approximately two days. Shorter sequence-length experiments are 10-20x faster, and can be run on a single GPU.

\subsection{Initial experiments}

We conducted two preliminary experiments to verify (i) the need for contextual training strategy and (ii) the need for in-batch false negative filtering when doing adversarial contrastive learning on a real dataset.

\begin{figure}[h]
    \centering
    \includegraphics[width=0.9\textwidth]{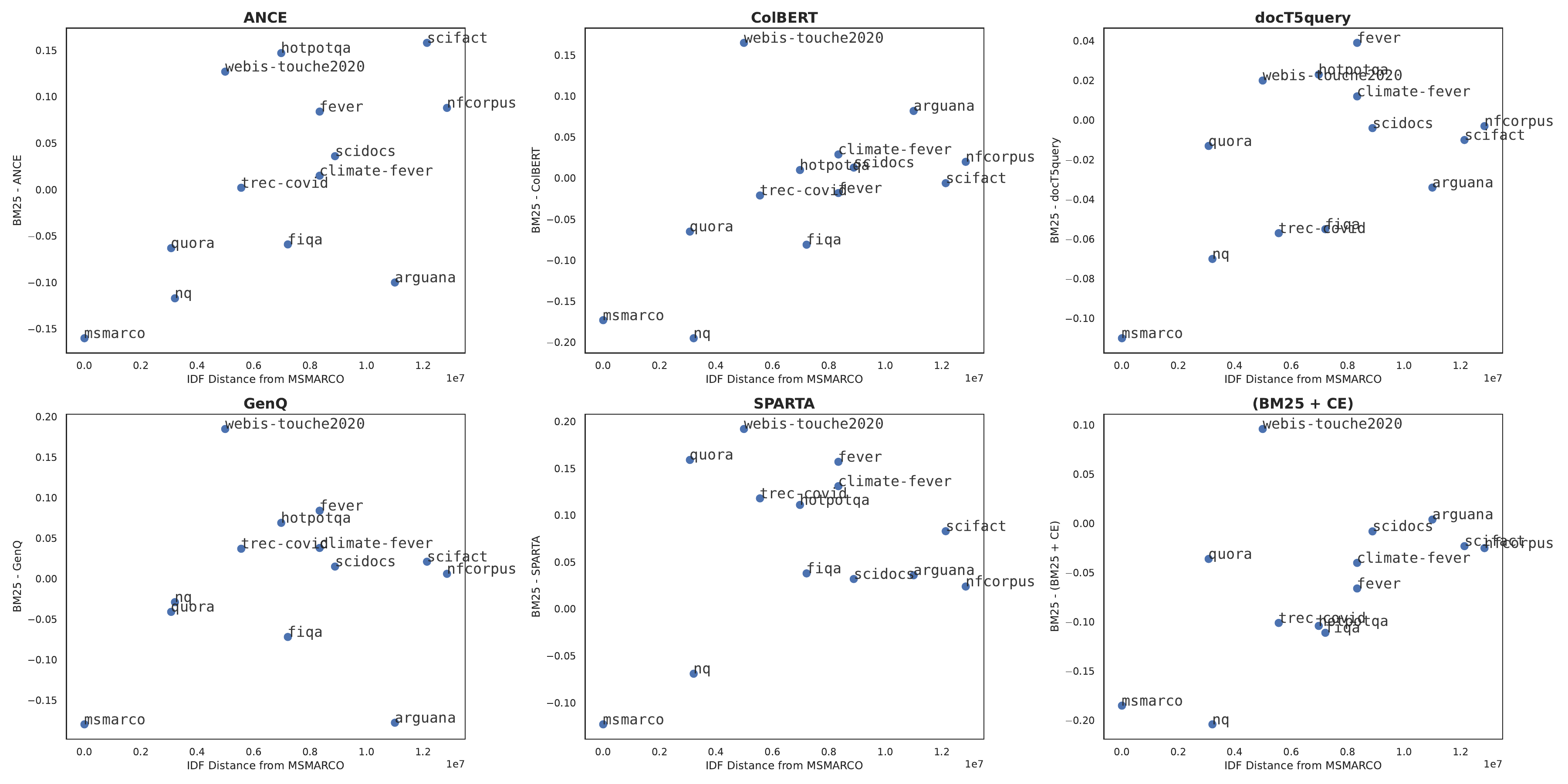}
    \caption{ Analysis of domain shift for popular neural retrieval methods. Performance difference from BM25 (y-axis) correlates with the different in IDF of the test corpus $\cal  D$ form the training corpus ${\cal D}_T$.}
    \label{fig:beir-bm25}
\end{figure}

\paragraph{Preliminary experiment (i). }We conduct a preliminary experiment to verify this issue. Starting from several trained retrieval systems we compute performance on a variety of different tasks from the BEIR dataset. Additionally we compute the IDF statistics from the datasets, and compare the divergence from the base IDF statistics of the training set. Figure~\ref{fig:beir-bm25} shows that datasets with high-divergence have very high correlation with the accuracy degradation of models when measured in comparison to BM25, which is able to measure and adapt to statistics of the test corpus. 

\paragraph{Preliminary experiment (ii).} \Jack{Provide context or split these up better} We select a random document from an unsupervised corpus and look at its nearest neighbors, displayed in \Cref{tab:cluster-neighbors}. We observe that the nearest neighbors to a given document in a large corpus are very close; in fact, many of them could be considered valid documents for the given query as well.   

\begin{table}[th]
  \caption{Nearest-neighbors to a single query in a large unsupervised dataset.}
  \label{tab:cluster-neighbors}
  \begin{tabular}{p{0.5\textwidth}p{0.5\textwidth}}
    \toprule
    Query & Document \\
    \midrule
    \addlinespace
    looks like my card payment was duplicated after all. [...] \\
    \midrule
    \addlinespace
    why is there an extra €1 fee in my statement? & why is there an extra charge on my statement? \\
    \addlinespace
    what is this fee for card payment? & why was a fee charged for my card payment? \\
    \addlinespace
    why do i have duplicate transactions for one purchase? & why was my transaction charged twice? \\
    \addlinespace
    i have two of the same charges on my account! & why was my transaction charged twice? \\
    \addlinespace
    my transaction went through but i was charged a fee. why? & why was a fee charged for my transfer? \\
    \addlinespace
    my account shows i have been charged twice for the same meal. [...] \\
    \addlinespace
    will i get extra charges? & why was a fee charged for my transfer? \\
    \addlinespace
    i got charged in double and want a refund & why was my transaction charged twice? \\
    \addlinespace
    where do i pay with my debit or credit card? & why is my card not accepted? \\
    \addlinespace
    why did i get charged a fee for my card payment? & why was a fee charged for my card payment? \\
    \addlinespace
    my statement shows different transaction times. & why was my transaction charged twice? \\
    \bottomrule
  \end{tabular}
\end{table}

This challenge motivates our embedding contextualization. In this section, we describe two complementary methods for remediation, (a) a contextual training method, (b) a contextual encoding method. 

\subsection{Interactions between Contrastive Loss and Distributed Data Parallel}

The authors note that it can be notoriously difficult to train models using both contrastive loss and the distributed data parallel (DDP) setting. In particular, when aggregating samples between GPUs, if any artifact reveals which GPU a model came from (for example, if the GPU model weights are initialized slightly differently) than the model can quickly deteriorate to a suboptimal solution, each GPU learning a different final model and ``cheating'' to classify samples based on which GPU they came from. 

This issue is made extra difficult by the fact that gradient-syncing must be disabled for large-batch contrastive learning to work efficiently. If gradient syncing becomes totally disabled, the training silently diverges as each model learns a degenerate solution. We advise practitioners to take care when controlling gradient-syncing and run many control experiments to determine performance equivalence between DDP and non-DDP scenarios.

One potential benefit of our method is that it greatly decreases the number of hard negatives required per batch, which means that negative-sharing across GPUs may not be necessary in most settings. If possible, the most sanity-preserving way to perform contrastive training could be to 

\subsection{Removing positionality with rotary embeddings}
\label{app:flash-attention-positionity-disabling}
One detail of our model architecture is that it does not track positionality between dataset input tokens. Although disabling positionality would be trivial an a BERT-like encoder model that uses learned positional embeddings, we use a version of BERT with \textit{rotary} positional embeddings which inject positional information at each layer of the transformer. To circumvent this step, we modify the model internals to set dataset input tokens to zero for the self-attention step only, and add a residual connection propagating the dataset input tokens past the attention phase.

\subsection{Additional results}

\begin{figure}[t]
    \centering
    \begin{minipage}{\textwidth}
        \begin{minipage}{0.48\textwidth}
            \includegraphics[width=\textwidth]{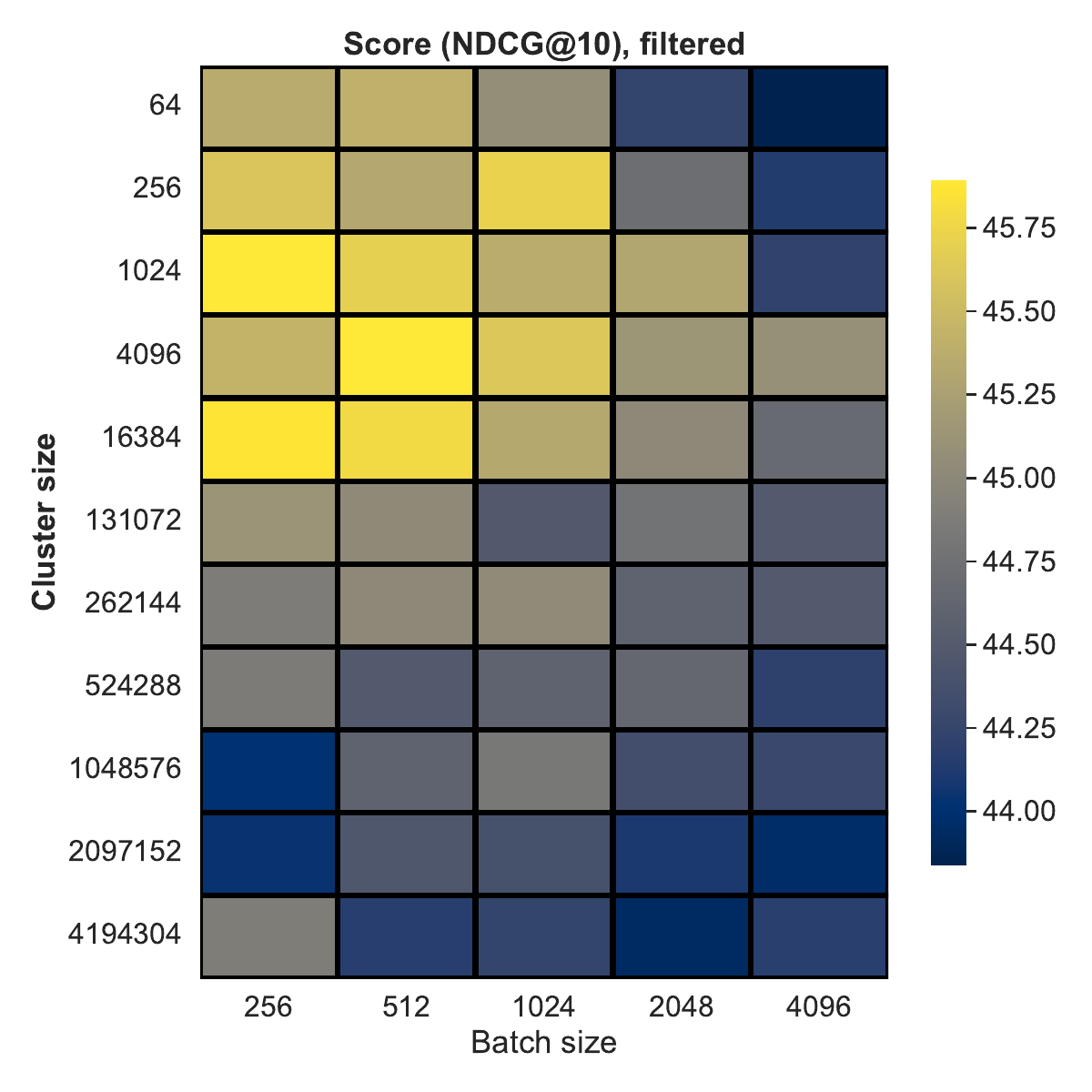}
        \end{minipage}
        \begin{minipage}{0.48\textwidth}
            \includegraphics[width=\textwidth]{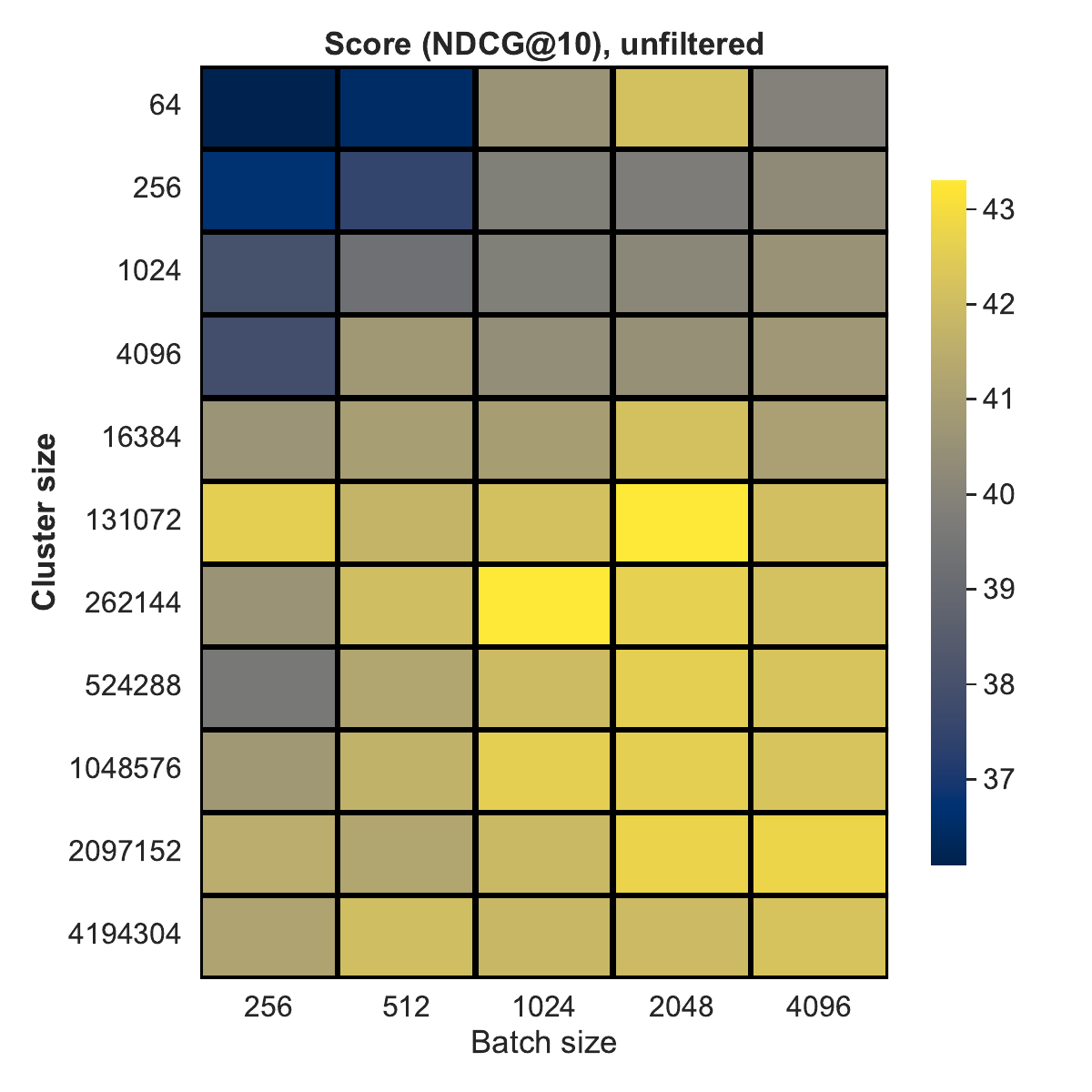}
        \end{minipage}
    \end{minipage}
    \label{fig:heatmap-contextual-filtered}
    \caption{Contextual performance with filtering (left) and without (right) across batch and cluster sizes during unsupervised contrastive pre-training. Here, clustering with small cluster sizes clearly improves performance, and larger batch sizes do not.}
\end{figure}

\Cref{fig:heatmap-contextual-filtered} show sweeps over batch and cluster sizes under our small experimental settings when performing unsupervised pretraining with contextual architecture. We see similar trends to those observed with the biencoder architecture, however we note that performance is higher across the board and our transductive model is able to perform well even at higher cluster sizes and low batch sizes.

One confounding factor in these experiments is that since the number of contextual documents is fixed, the number of different contextual inputs seen during training decreases with higher batch size. This might explain part of why performance stagnates with higher batch sizes; increasing the batch size decreases the total number of learning examples seen by our contextual model.

\begin{figure}[t]
    \includegraphics[width=\textwidth]{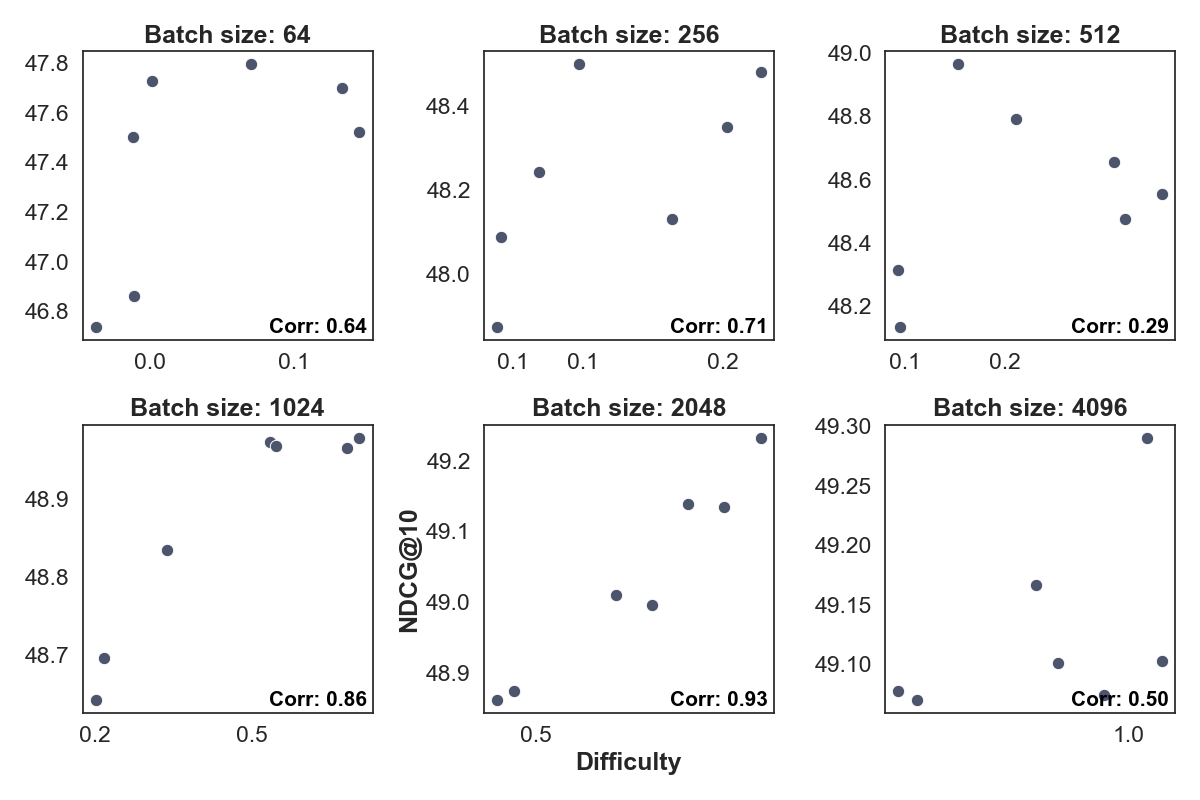}
    \label{fig:cluster-hardness-supervised}
    \caption{Correlation between batch difficulty and perforamnce after supervised training.}
\end{figure}

\paragraph{Supervised training: difficulty correlations.} In \Cref{fig:cluster-hardness-supervised} we plot the correlation between batch difficulty and downstream performance across cluster sizes (and within batch sizes) in the supervised setting. In this case we also see the best performance through the most difficult clusters.

\begin{figure}[t]
    \centering
    \includegraphics[width=.5\textwidth]{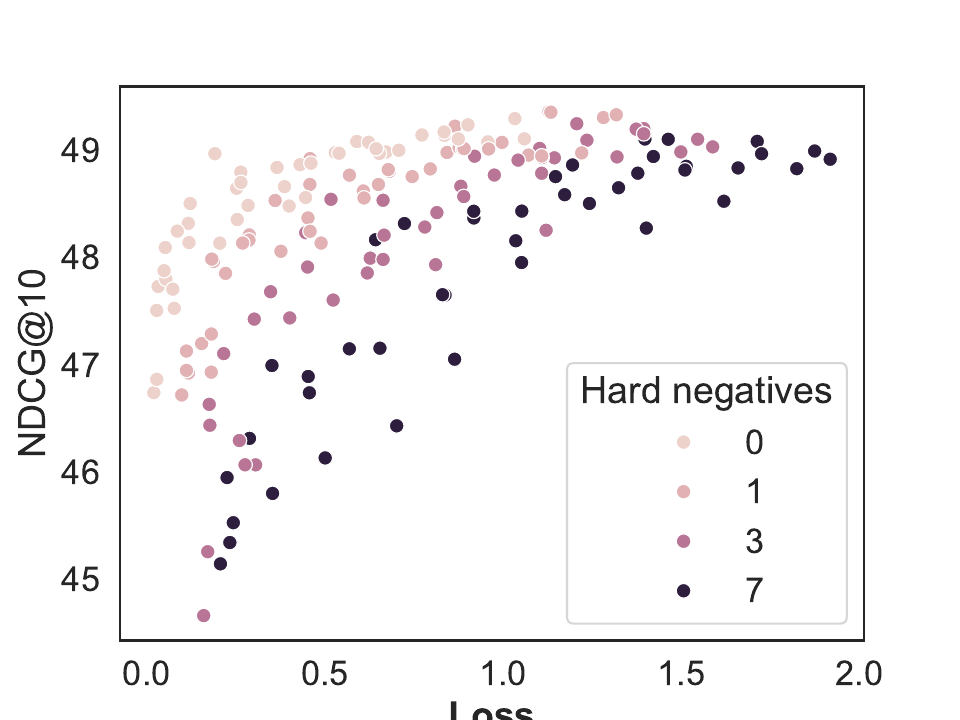}
    \label{fig:cluster-perforamnce-supervised}
    \caption{Performance of all supervised models, across numbers of hard negatives.}
\end{figure}

\paragraph{Supervised training: full results.} We plot the full results of all supervised training experiments in \Cref{fig:cluster-perforamnce-supervised}. Our experiments in this setting (using the mined negatives from the Nomic supervised meta-datasets) generally show \textit{decreasing} performance with additional hard negatives.

\begin{figure}
    \centering
    \includegraphics[width=.5\textwidth]{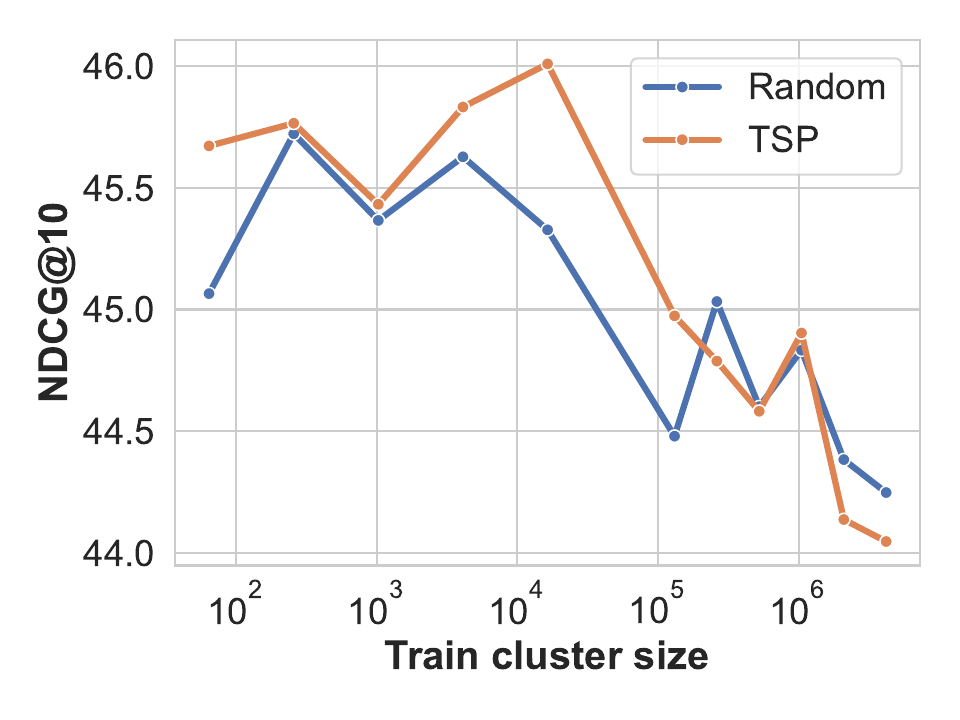}
    \caption{Pre-training with TSP vs. random batching across cluster sizes.}
    \label{fig:tsp-packing}
\end{figure}

\paragraph{TSP Packing.} We compare randomly packing clusters into batches vs. a greedy traveling salesman-style solution, similar to \citep{shi2024incontextpretraining}. In our scenario, we first cluster datapoints, then find the centroid embedding of each cluster. We begin packing by randomly selecting a cluster, and then choose the next cluster by finding the cluster with the closest centroid to the current one. Results are shown in \Cref{fig:tsp-packing}. Although these results appear slightly noisy, we see an improvement from TSP-style packing especially at smaller cluster sizes (where packing has an outsized impact). We therefore opt to use this packing procedure for our main model.

\begin{figure}
    \begin{minipage}{0.48\textwidth}
        \includegraphics[width=\textwidth]{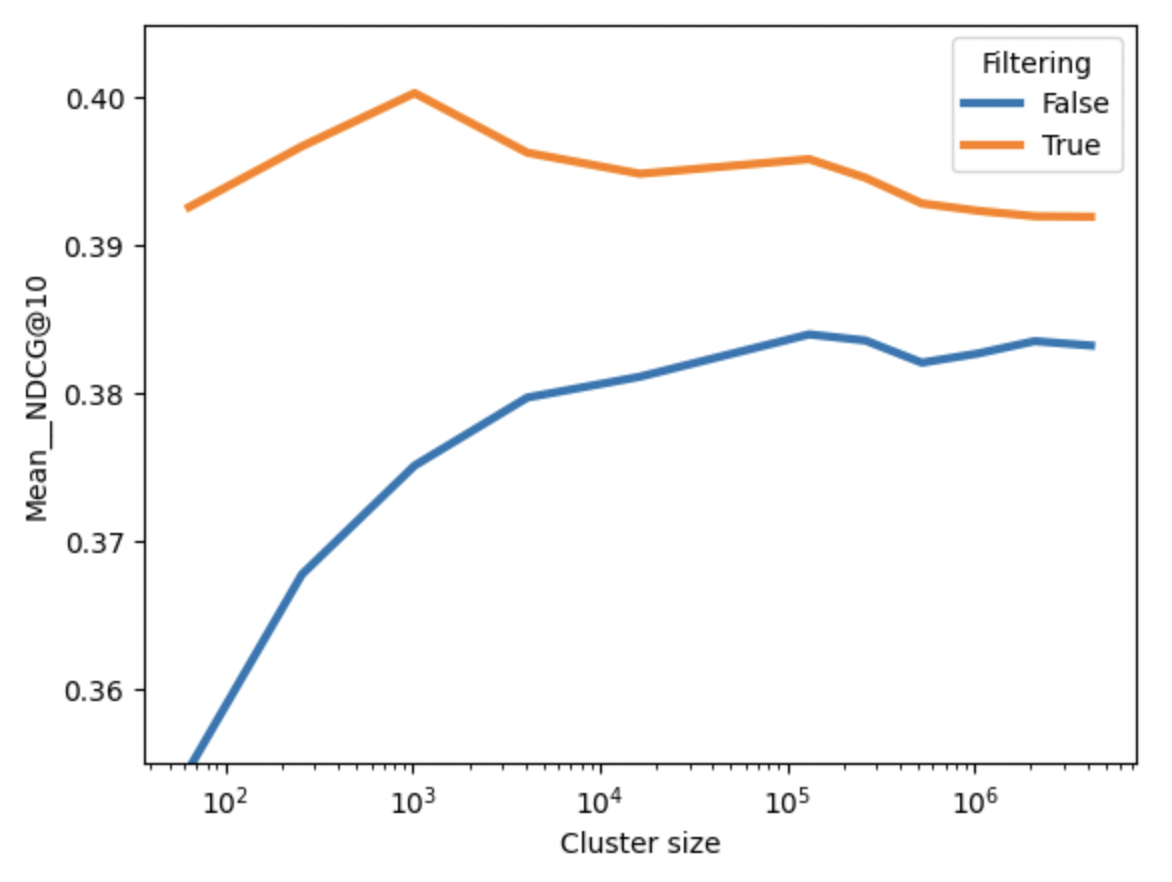}
        \caption{Model performance vs. cluster size with and without filtering. When false negative filtering is enabled, we see more improvements in performance from clustering at small cluster sizes.}
        \label{fig:line-cluster-size}
    \end{minipage}
    \hspace{0.05\textwidth}
    \begin{minipage}{0.48\textwidth}
        \includegraphics[width=\textwidth]{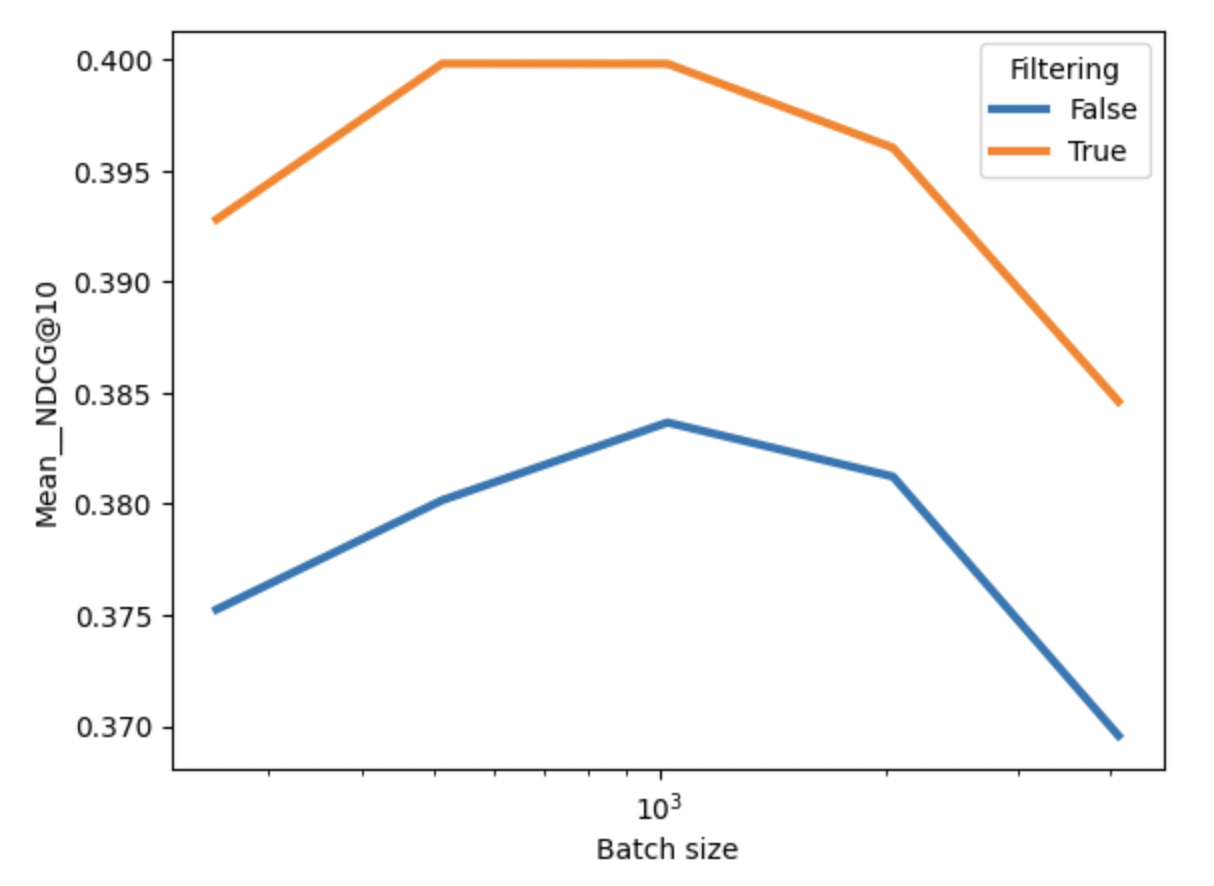}
        \caption{Model performance vs. batch size with and without filtering. With and without filtering, the optimal batch size ranges between $10^2$ and $10^4$; performance starts to decrease as batch size grows too large.}
        \label{fig:line-batch-size}
    \end{minipage}
\end{figure}

\begin{figure}[t]
    \centering
    \begin{minipage}[b]{0.45\textwidth}
        \centering
        \includegraphics[width=\textwidth, trim={1cm 0.7cm 1cm 4cm},clip]{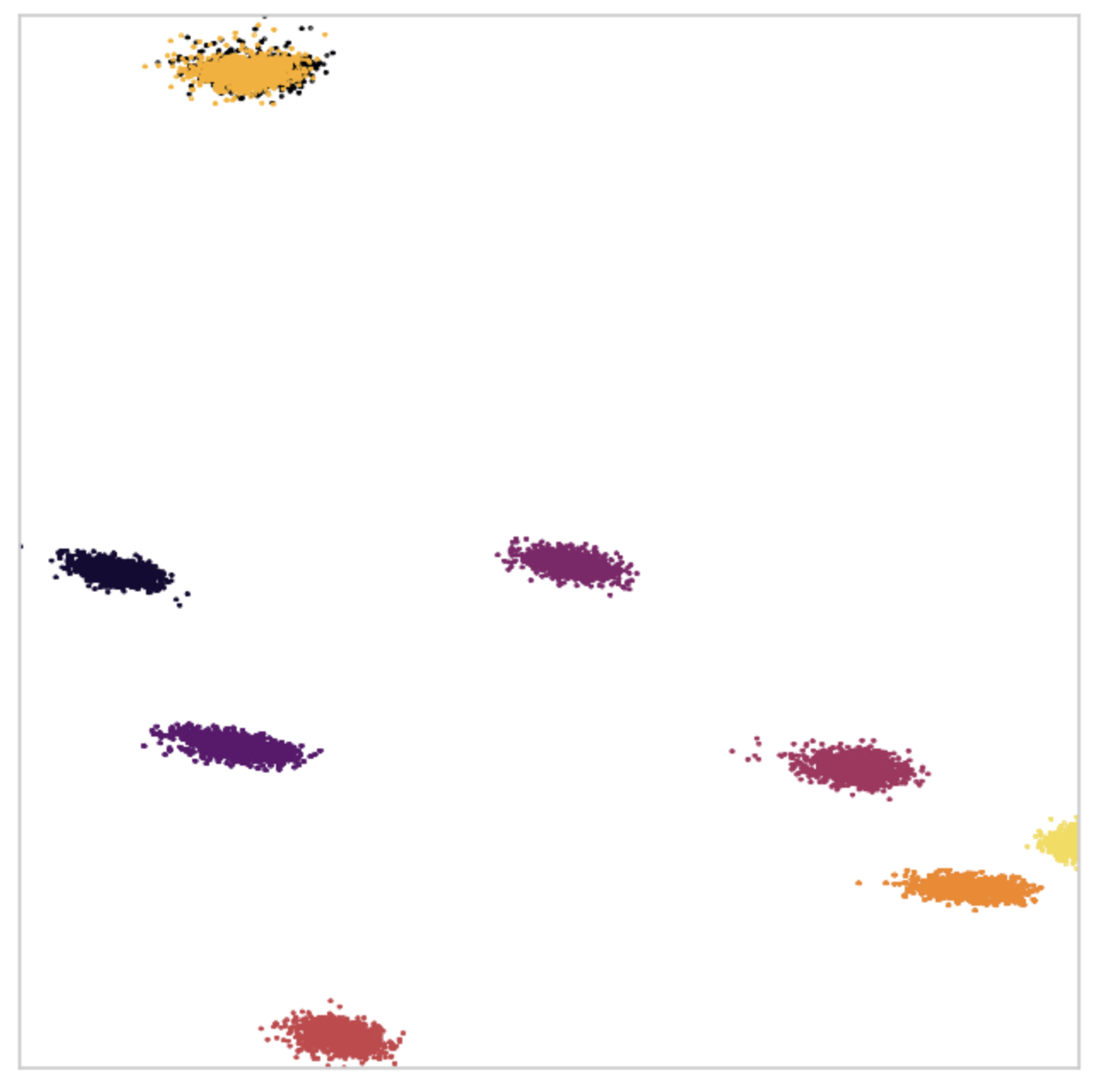}
        \label{fig:anchor}
        \caption{Each color indicates a single document input $d$. Different points represent different values $\phi(d; {\cal D})$ for different contexts.} 
    \end{minipage}
    \hfill
    \begin{minipage}[b]{0.45\textwidth}
        \centering
        \includegraphics[width=\textwidth]{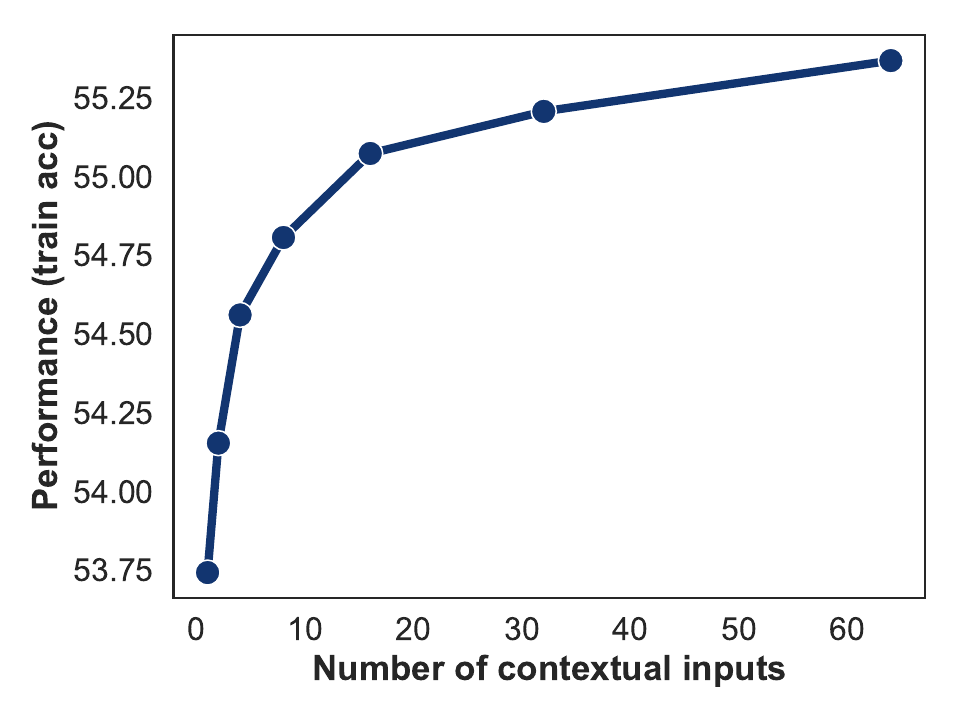}
        \label{fig:perf_vs_corpus}
        \caption{Performance of CDE model as the number of contextual examples increases.}   
    \end{minipage}
\end{figure}

\paragraph{Impact of context size} We consider contextual embeddings might move in space as their conditioning varies. \Cref{fig:anchor} displays a few qualitative examples. We generate embeddings for randomly sampled documents from the TREC-Covid dataset and visualize their embeddings with PCA, where unique document inputs with different contextual embeddings are visualized in the same color. By changing only the conditioning we reshape the embedding space and our model produces different embedding for the same text. Note that although the embeddings are clearly moving in response to changing the contextual inputs, they still remain closer to each other than to different documents.

We also consider how additional context is improving our model. Because the model includes an optional null token, we can supply any number of contextual inputs. We plot our model's performance across context sizes in Figure \ref{fig:perf_vs_corpus}. We see that our model is able to utilize partial context window sizes, and even perform reasonably with no context (i.e. all null token inputs) but offers the best performance given a full context window size.

\subsection{Cluster text examples}

\begin{table}[]
\footnotesize
\begin{tabularx}{\textwidth}{|X|X|}
\toprule
query & document \\
\midrule 
population of breckenridge mi & breckenridge, michigan. breckenridge is a village in gratiot county in the u. s. state of michigan. the population was 1, 328 at the 2010 census. the village is located in wheeler township. \\
can a deposition be used in a criminal case & depositions are commonly used in civil litigation (suits for money damages or equitable relief) [...] \\
what cases require strict scrutiny & the strict scrutiny standard is one of three employed by the courts in reviewing laws and government policies. the rational basis [...] \\
function of state supreme courts & it has also initiated several programs designed to improve the effectiveness of the court system. a primary function of the supreme court is to ensure [...] \\
what is the population in idaho & idaho ’ s population grows to nearly 1. 7 million. idaho ’ s population grew by 1. 2 percent between mid - 2014 and mid - 2015, the 12th strongest increase among the states and four - tenths of a percentage point ahead of the national growth rate. \\
what is the population of manson, ia & manson, iowa. manson is a city in calhoun county, iowa, united states. the population was 1, 690 at the 2010 census. \\
what happens after a sentencing hearing & find answers. sentencing. after a criminal defendant is convicted or pleads guilty, a judge will decide [...] \\
flathead county population & flathead county, montana. flathead county is a county located in the u. s. state of montana. as of the 2010 census, the population was 90, 928, making it [...] \\
whiting, ks population & the city of whiting had a population of 177 as of july 1, 2017. whiting ranks in the lower quartile for population density and diversity index when compared to the other cities, towns [...] \\
what is the population of lewiston id & lewiston, id population and races. as of 2010 - 2014, the total population of lewiston is 32, 178, which is 4. 12\% more than it was in 2000. [...] \\
what happens if you don't show up for jury & what happens if you don't show up for jury duty in california? a : according to california courts, judicial branch of california, if a citizen fails to show up for jury duty, the juror can accrue fines up to \$1,500. if service presents an undue hardship, a juror can request a postponement or to be excused. otherwise, citizens are not exempt from jury duty. \\
population of clearfield county pa & clearfield is a borough and the county seat of clearfield county, pennsylvania, united states. the population was 6, 215 at the 2010 census, and the borough is part of the dubois, pa micropolitan statistical area, as well as the larger state college - dubois, pa combined statistical area. \\
how long can it take for a trial & the preliminary hearing phase of the trial usually takes place 5 - 6 days after an arraignment. in the case of a misdemeanor [...] \\
population clinton ky & clinton county is a county located in the u. s. state of kentucky. as of the 2010 census, the population was 10, 272. its county seat is albany. the county was formed in 1835 and named for dewitt clinton, the seventh governor of new york. it is a prohibition or dry county. \\
population of iosco county michigan & with 25, 420 people, iosco county is the 55th most populated county in the state of michigan out of 83 counties. but watch out, iosco county, because gladwin county with 25, 411 people and manistee county with 24, 420 people are right behind you. \\ \bottomrule
\end{tabularx}
\caption{Sixteen samples from a cluster our algorithm finds in the supervised training data. The full cluster size is $256$ points out of a dataset of $1.5M$.}
\label{tab:app-cluster-ex}
\end{table}

We include random examples from a cluster gathered from our supervised dataset, shown in \Cref{tab:app-cluster-ex}. This particular cluster appears to be a combination of documents about county populations in the Untied States (in Kentucky, Iowa, Pennsylvania, etc.) and documents about criminal trials (mentioning hearings, depositions, and courts).

\subsection{Task prefixes}
\label{app:task-prefixes}

Prefixes are hand-written for each dataset in both meta-training sets. We follow the same prefix selection procedure as \citet{nussbaum2024nomic}, inspired by \citet{reimers2023cohereembed}:

\begin{itemize}
    \item \texttt{search\_query}
    \item \texttt{search\_document}
    \item \texttt{classification}
    \item \texttt{clustering}
\end{itemize}

\subsection{Unsupervised training datasets}

\begin{table*}
   \centering
   \caption{Distribution of pretraining datasets curated in \citet{nussbaum2024nomic}.}
    \label{tab:unsupervised-datasets}
\begin{minipage}{\textwidth}
\begin{tabularx}{\textwidth}{lXX}
  \toprule
  Dataset & Datapoints & $\%$ Dataset  \\
  \midrule
  Reddit\footnote{\url{https://huggingface.co/datasets/sentence-transformers/reddit-title-body}} & 64,978,944 & 0.28 \\
  PAQ \cite{lewis2021paq} & 52,953,088 & 0.23 \\
  Amazon Reviews \cite{ni-etal-2019-justifying} & 38,682,624 & 0.16 \\
  S2ORC Title Abstract \cite{lo2020s2orc} & 35438592 & 0.15 \\
  WikiAnswers \cite{fader2014openqa} & 9,912,320 & 0.04 \\
  S2ORC Citation Titles \cite{lo2020s2orc} & 7,585,792 & 0.03 \\
  S2ORC Abstract Citation \cite{lo2020s2orc} & 7,503,872 & 0.03 \\
  S2ORC Abstract Body \cite{lo2020s2orc} & 6,389,760 & 0.03 \\
  Wikipedia Title Body \cite{wikidump2024} & 6,078,464 & 0.03 \\
  Gooaq \cite{khashabi2021gooaq} & 1,245,184 & 0.01 \\
  Codesearch \cite{husain2019codesearchnet} & 835,584 & $<$.01 \\
  AGNews \cite{zhang2016characterlevel} & 409,600 & $<$.01 \\
  CCNews \cite{hamborg2017agnews} & 344,064 & $<$.01 \\
  NPR \footnote{\url{https://files.pushshift.io/news/}} & 344,064 & $<$.01 \\
  CNN \cite{see2017npr} & 278,528 & $<$.01 \\
  Yahoo Title-Answer \footnote{\url{https://www.kaggle.com/soumikrakshit/yahoo-answers-dataset}} & 262,144 & $<$.01 \\
  AmazonQA \cite{gupta2019amazonqa} & 212,992 & $<$.01 \\
  Yahoo Title-Question \footnote{\url{https://www.kaggle.com/soumikrakshit/yahoo-answers-dataset}} & 196,608 & $<$.01 \\
  Sentence Compression \cite{filippova2013overcoming} & 163,840 & $<$.01 \\
  YahooQA \footnote{\url{https://www.kaggle.com/soumikrakshit/yahoo-answers-dataset}} & 131,072 & $<$.01 \\
  ELI5 \cite{fan2019eli5} & 98,304 & $<$.01 \\
  Altlex \cite{hidey2016identifying} & 98,304 & $<$.01 \\
  Wikihow \cite{koupaee2018wikihow} & 81,920 & $<$.01 \\
  SimpleWiki \cite{coster2011simplewiki} & 81,920 & $<$.01 \\
  StackExchange Duplicate Questions \footnote{\url{https://data.stackexchange.com/apple/query/fork/1456963}} & 65,536 & $<$.01 \\
  StackExchange Title Body \footnote{\url{https://data.stackexchange.com/apple/query/fork/1456963}} & 65,536 & $<$.01 \\
  StackExchange Body Body \footnote{\url{https://data.stackexchange.com/apple/query/fork/1456963}} & 65,536 & $<$.01 \\
  Quora Duplicate Questions \footnote{\url{https://quoradata.quora.com/First-Quora-Dataset-Release-Question-Pairs}} & 32,768 & $<$.01 \\
  SQuAD \cite{rajpurkar2016squad} & 16,384 & $<$.01 \\
  \bottomrule
  Total & 234,553,344 & 1 
\end{tabularx}
\end{minipage}
\end{table*}

We train on $234M$ weakly supervised query-document pairs collected for training text embedding models in \citet{nussbaum2024nomic}. The full distribution of $29$ datasets is shown in \Cref{tab:unsupervised-datasets}. Reddit alone makes up over 25\% of the data distribution, with $19$ of the datasets comprising under 1\% of the total data.

\subsection{BEIR evaluation datasets}

Our initial experiments involve evaluating on nine datasets from the BEIR benchmark. Datasets are detailed in \Cref{tab:beir-eval-datasets}. To enable fast evaluation at this stage, we obtain the top $1024$ relevant documents to each document with GTR \citep{ni2021gtr} and rerank only these documents at evaluation time.

\begin{table}
   \centering
   \caption{Distribution of BEIR evaluation datasets used, ordered by corpus size.}
   \label{tab:beir-eval-datasets}
   \begin{tabular}{lcc}
      \toprule
      Dataset & Queries & Documents \\
      \midrule
      NFCorpus & 323 & 3,633 \\
      SciFact & 300 & 5,183 \\
      ArguAna & 1,406 & 8,674 \\
      SciDocs & 1,000 & 25,657 \\
      TREC-COVID & 50 & 171,332 \\
      Quora & 5,000 & 522,931 \\
      Natural Questions & 3,452 & 2,681,468 \\
      MS MARCO & 6,980 & 8,841,823 \\
      \bottomrule
   \end{tabular}
\end{table}

\subsection{Additional modeling ablations}

\begin{figure}
    \centering
    \includegraphics[width=.5\textwidth]{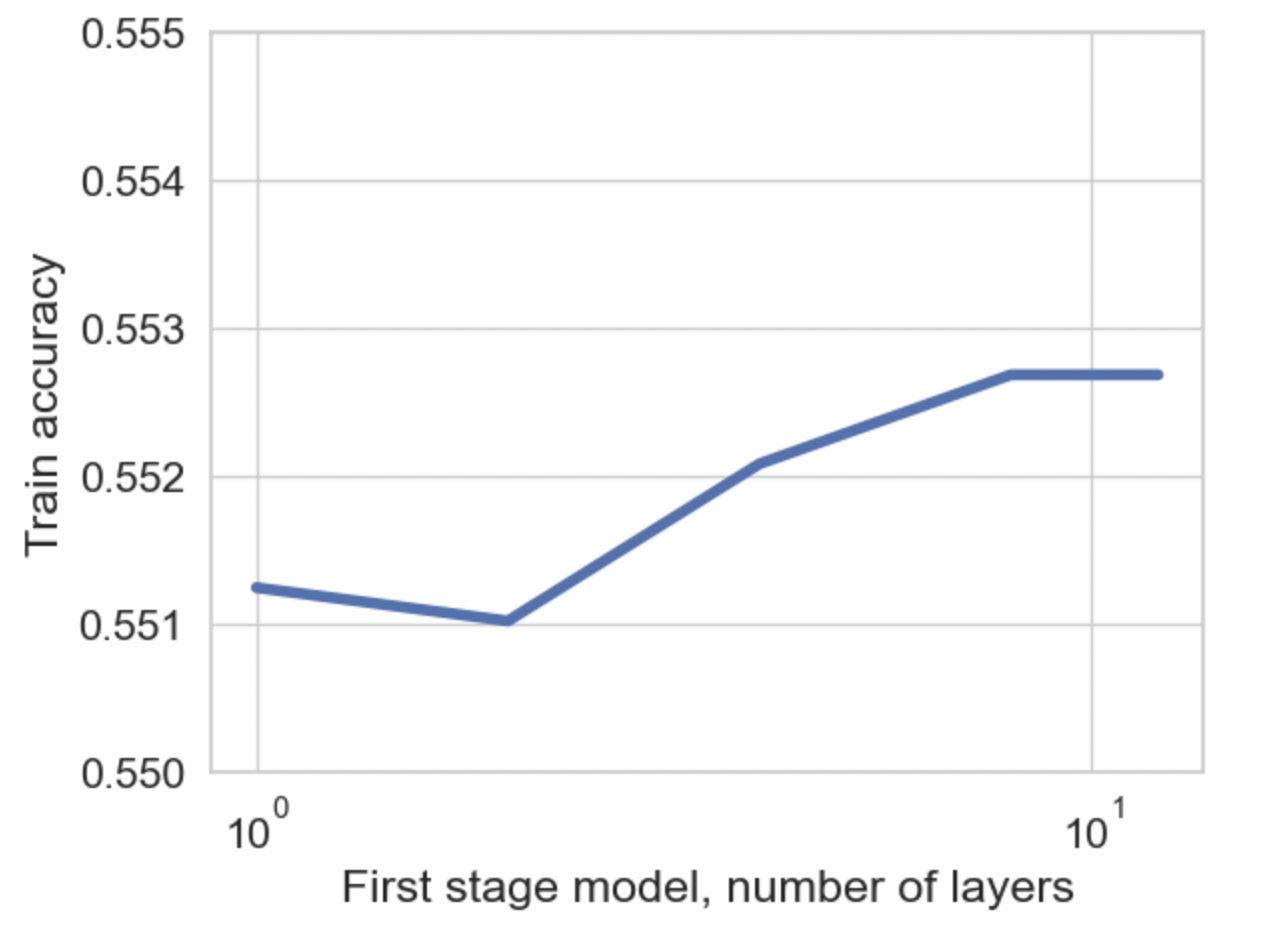}
    \label{fig:app-scaling-first-stage}
    \caption{System performance (training accuracy) as we scale the size of the first-stage model encoder only.}   
\end{figure}

\textbf{First-stage model size.} One consideration is whether we can improve our system without affecting search inference time by scaling the number of parameters in the backbone model only. We study this affect by scaling the number of layers in the transformer backbone of the first-stage model from $1$ to the full $12$. Resulting performance is shown in \Cref{fig:app-scaling-first-stage}.

Our results show that scaling the first-stage model has a small positive influence on model performance. However, since the total improvement from a $12$x increase in first-stage model size is less than one percent, we conclude that the second-stage model size has a much larger impact on performance.

\subsection{How many tokens per document?}

\begin{figure}
    \centering
    \includegraphics[width=\textwidth]{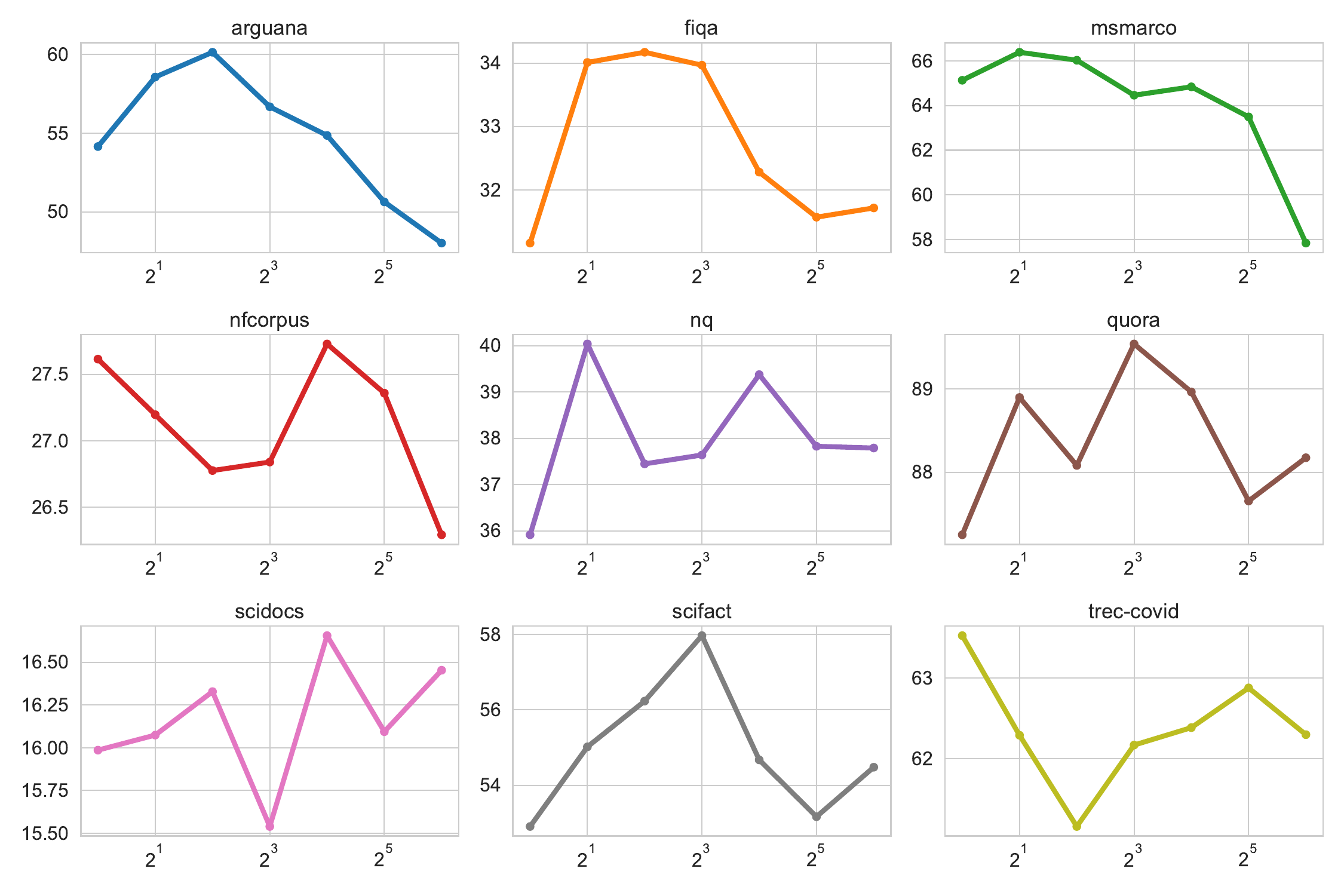}
    \label{fig:app-tokens-per-doc}
    \caption{Performance per-dataset as we scale tokens-per-document, while keeping the total number of contextual tokens fixed. Different domains prefer a different number of tokens per document.}   
\end{figure}

We consider the question of how many tokens per document is ideal while keeping the total number of document tokens fixed. Results per the nine evaluation datasets of BEIR are shown in \Cref{fig:app-tokens-per-doc}.

\subsection{MTEB Retrieval Evaluation Performance}

\begin{table}[ht]
  \centering
  \tiny
  \begin{tabular}{lcccccccccccccccc}
    \toprule
    Method & Arg & CQA & CFEVER & DBP & FEVER & FiQA & HPQA & MSMRC & NFC & NQ & QUORA & SCID & SCIF & TREC & TOUCHE & Mean \\
    \midrule
    \textbf{Unsupervised} \\
    Baseline & 54.8 & 41.4 & 24.7 & 40.2 & 74.4 & 39.9 & 63.8 & 35.0 & 35.7 & 48.6 & 88.2 & 20.2 & 72.0 & 62.2 & 19.2 & 48.0 \\
    Contextual & 54.9 & 43.1 & 24.4 & 40.7 & 79.6 & 42.1 & 68.8 & 38.9 & 36.5 & 57.8 & 88.9 & 21.1 & 72.8 & 77.1 & 21.9 & 51.2 \\
    \midrule
    \textbf{Supervised} \\
    Baseline & 49.3 & 40.5 & 38.3 & 45.0 & 85.0 & 38.4 & 73.6 & 43.1 & 35.0 & 59.4 & 87.7 & 18.3 & 70.5 & 79.9 & 28.2 & 52.8 \\
    Contextual & 53.8 & 41.2 & 38.8 & 43.3 & 89.2 & 40.1 & 73.9 & 42.2 & 35.9 & 61.6 & 87.1 & 20.1 & 72.7 & 82.6 & 27.8 & 54.0 \\
    \bottomrule
    \end{tabular}
  \caption{Results (NDCG@10) on the retrieval setting of the MTEB benchmark.}
  \label{tab:mteb-retrieval}
\end{table}

To evaluate on MTEB, we subsample contextual documents from the full corpus available in each dataset and modality. For retrieval, this corresponds to the corpus itself (importantly, not the queries); for other modalities, we choose the default ``text'' field in each casel. For classification tasks, we sample from the text side (not the classification labels themselves).

\Cref{tab:mteb-retrieval} shows our model performance on all datasets in the MTEB retrieval category. We see largest improvements over the baseline on the ArguAna and TREC-Covid datasets.

\end{document}